\def\year{2019}\relax
\documentclass[letterpaper]{article} 
\usepackage{aaai19}  
\usepackage{times}  
\usepackage{helvet}  
\usepackage{courier}  
\usepackage{url}  
\usepackage{graphicx}  

\usepackage{amsfonts}       
\usepackage{nicefrac}       
\usepackage{microtype}      
\usepackage{amsmath}
\usepackage{amssymb}
\usepackage{dsfont}

\usepackage[font=small,labelfont=sf,skip=0pt]{caption}
\usepackage[caption=true,font=small,labelfont=sf,textfont=sf,skip=0pt]{subfig}
\usepackage{wrapfig}
\usepackage{floatrow}
\usepackage{color}
\usepackage{algorithmic}
\usepackage{algorithm}
\usepackage[compact]{titlesec}
\usepackage{amsthm}

\DeclareMathOperator*{\argmin}{arg\,min}
\DeclareMathOperator*{\argmax}{arg\,max}

\begin{document}
%
\title{Active Anomaly Detection via Ensembles}
\author{Shubhomoy Das, Md Rakibul Islam, Nitthilan Kannappan Jayakodi, Janardhan Rao Doppa\\
	\texttt{\{shubhomoy.das,mdrakibul.islam,n.kannappanjayakodi,jana.doppa\}@wsu.edu}\\
	Washington State University\\
	Pullman, WA 99163\\
}
\maketitle
\begin{abstract}
In critical applications of anomaly detection including computer security and fraud prevention, the anomaly detector must be configurable by the analyst to minimize the effort on false positives. One important way to configure the anomaly detector is by providing true labels for a few instances. We study the problem of label-efficient active learning to automatically tune anomaly detection ensembles and make four main contributions. First, we present an important insight into how anomaly detector ensembles are naturally suited for active learning. This insight allows us to relate the greedy querying strategy to uncertainty sampling, with implications for label-efficiency. Second, we present a novel formalism called {\em compact description} to describe the discovered anomalies and show that it can also be employed to improve the diversity of the instances presented to the analyst without loss in the anomaly discovery rate. Third, we present a novel data drift detection algorithm that not only detects the drift robustly, but also allows us to take corrective actions to adapt the detector in a principled manner. Fourth, we present extensive experiments to evaluate our insights and algorithms in both batch and streaming settings. Our results show that in addition to discovering significantly more anomalies than state-of-the-art unsupervised baselines, our active learning algorithms under the streaming-data setup are competitive with the batch setup.
\end{abstract}

\section{Introduction}

We consider the problem of {\em anomaly detection}, where the goal is to detect unusual but interesting data (referred to as {\em anomalies}) among the regular data (referred to as {\em nominals}). This problem has many real-world applications including credit card transactions, medical diagnostics, computer security, etc., where anomalies point to the presence of phenomena such as fraud, disease, malpractices, or novelty which could have a large impact on the domain, making their timely detection crucial.

Anomaly detection poses severe challenges that are not seen in traditional learning problems. First, anomalies are significantly fewer in number than nominal instances. Second, unlike classification problems, no hard decision boundary exists to separate anomalies and nominals. Instead, anomaly detection algorithms train models to compute scores for all instances, and report instances which receive the highest scores as anomalies. Most of these algorithms \cite{chandola:09} only report technical outliers, i.e., data which do not fit a \textit{normal} model as anomalies. Such outliers are not guaranteed to be interesting and result in many false positives when they are not. 

Prior work on anomaly detection has three main shortcomings: 1) Many algorithms are unsupervised in nature and do not provide a way to configure the anomaly detector by the human analyst to minimize the effort on false-positives. There is little to no work on principled active learning algorithms; 2) There is very little work on enhancing the diversity of discovered anomalies \cite{gornitz:2013}; and 3) Most algorithms are designed to handle batch data well, but there are few principled algorithms to handle streaming data setting.


{\bf Contributions.} We study label-efficient active learning algorithms to improve unsupervised anomaly detector ensembles and address the above three shortcomings of prior work in a principled manner: {\bf (1)} We present an important insight into how anomaly detector ensembles are naturally suited for active learning, and why the greedy querying strategy of seeking labels for instances with the highest anomaly scores is efficient; {\bf (2)} We present a novel approach to describe the discovered anomalies and show that it can also be employed to improve the diversity of the instances presented to the analyst while achieving an anomaly discovery rate comparable to the greedy strategy; {\bf (3)} We present a novel algorithm to robustly detect drift in data streams and to adapt the detector in a principled manner; and {\bf (4)} We present extensive empirical evidence in support of our insights and algorithms in both batch and streaming settings. We also demonstrate the improvement in diversity through an objective measure relevant to real-world settings. Our results show that in addition to discovering significantly more anomalies than state-of-the-art unsupervised baselines, our active learning algorithm under the streaming-data setup is competitive with the batch setup. Our code and data are publicly available\footnote{\scriptsize{\url{https://github.com/shubhomoydas/ad_examples}}}.

	

\section{Related Work}
\label{sec:related-work}

\noindent \textbf{Unsupervised anomaly detection algorithms} are trained without labeled data, and have assumptions baked into the model about what defines an anomaly or a nominal \cite{breunig:00,liu:08,pevny:2015,emmott:2015}. They typically cannot change their behavior to correct for the false positives after they have been deployed. Ensembles of unsupervised anomaly detectors \cite{aggarwal:2017,chen:2017,rayana:2015,pevny:2015,emmott:2015,senator:2013,liu:08,lazarevic:05} try to guard against the bias induced by a single detector by incorporating decisions from multiple detectors. 


\noindent \textbf{Active learning} corresponds to the setup where the learning algorithm can selectively query a human analyst for labels of input instances to improve its prediction accuracy. The overall goal is to minimize the number of queries to reach the target performance. There is a significant body of work on both theoretical analysis \cite{freund:1997,balcan:2007,balcan:2015,monteleoni:2006,dasgupta:2009,yan:2017} and applications \cite{settles:2010} of active learning.

\noindent \textbf{Active learning for anomaly detection} has recently gained prominence \cite{veeramachaneni:2016,das:2016,guha:2016,nissim:14,stokes:2008,he:2008} due to significant rise in the volume of data in real-world settings, which made reduction in false-positives much more critical. In this setting, the human analyst provides feedback to the algorithm on true labels. 
If the algorithm makes wrong predictions, it updates its model parameters to be consistent with the analyst's feedback. Some of these methods are based on ensembles and support streaming data \cite{veeramachaneni:2016,stokes:2008} but maintain separate models for anomalies and nominals internally, and do not exploit the inherent strength of the ensemble in a systematic manner.


\section{Anomaly Detection Ensembles}

\noindent {\bf Problem Setup.} We are given a dataset ${\bf D} = \{{\bf x}_1, ..., {\bf x}_n\}$, where ${\bf x}_i \in \mathds{R}^d$ is a data instance that is associated with a hidden label $y_i \in \{-1, +1\}$. Instances labeled $+1$ represent the \textit{anomaly} class and are at most a small fraction $\tau$ of all instances. The label $-1$ represents the \textit{nominal} class. We also assume the availability of an ensemble $\mathcal{E}$ of $m$ anomaly detectors which assigns scores ${\bf z} = \{z^1, ..., z^m\}$ to each instance ${\bf x}$ such that instances labeled $+1$ tend to have scores higher than instances labeled $-1$. We denote the ensemble score matrix for all unlabeled instances by ${\mathbf H}$. The score matrix for the set of instances labeled $+1$ is denoted by ${\mathbf H}_+$, and the matrix for those labeled $-1$ is denoted by ${\mathbf H}_-$. We assume a linear model with weights ${\mathbf w} \in \mathds{R}^m$ that will be used to combine the scores of $m$ anomaly detectors as follows: \texttt{Score}(${\bf x}$) = ${\mathbf w} \cdot {\bf z}$, where ${\bf z} \in \mathds{R}^m$ correspond to the scores from anomaly detectors for instance ${\bf x}$. This linear hyperplane separates anomalies from nominals. We will denote the optimal weight vector by ${\bf w}^*$. Our active learning algorithm $\mathcal{A}$ assumes the availability of an analyst who can provide the true label for any instance. The goal of $\mathcal{A}$ is to learn optimal weights for maximizing the number of true anomalies shown to the analyst.

\subsection{Suitability of Ensembles for Active Learning}
\label{sec:insights}

In this section, we show how anomaly detection ensembles are naturally suited for active learning as motivated by the active learning theory for standard classification.

\begin{figure*}[t]
	\centering
	\subfloat[\textbf{C1}: Common case]{\includegraphics[width=1.2in,height=1.2in]{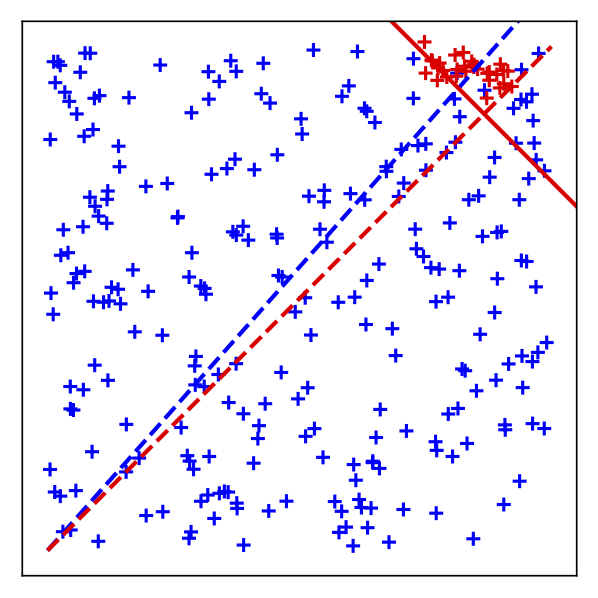} \label{fig:idealized_rect}}
	\subfloat[\textbf{C2}: Similar to Active Learning theory]{\includegraphics[width=2.6in,height=1.2in]{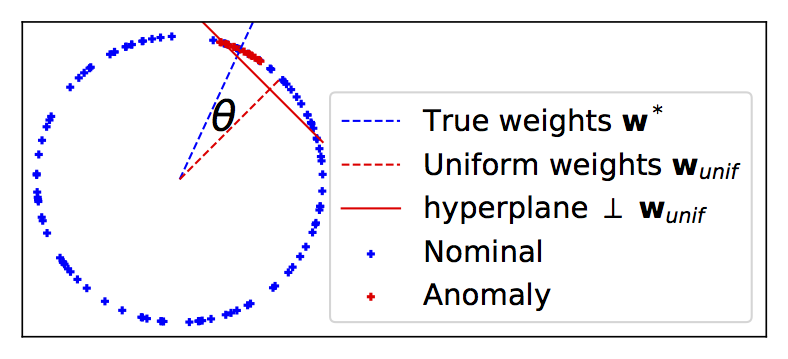} \label{fig:idealized_circ}}
	\subfloat[\textbf{C3}: IFOR case]{\includegraphics[width=1.2in,height=1.2in]{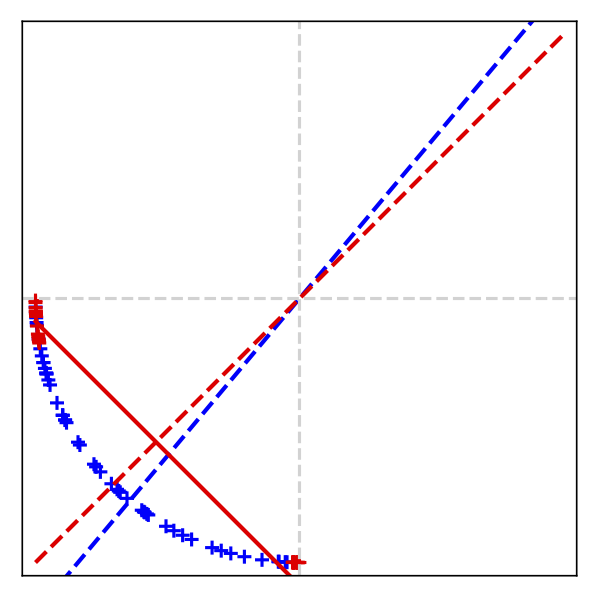} \label{fig:idealized_ifor}} \\[-1ex]
	\caption{Illustration of candidate score distributions from an ensemble in 2D. The two axes represent two different ensemble members.}
	\label{fig:idealized}
\end{figure*}

Without loss of generality, we assume that all scores from the members of an ensemble of anomaly detectors are normalized (i.e., they lie in $[-1, 1]$ or $[0, 1]$), with higher scores implying more anomalous. For the following discussion, ${\mathbf w}_{unif} \in \mathds{R}^m$ represents a vector of equal values, and $||{\mathbf w}_{unif}|| = 1$. Figure~\ref{fig:idealized} illustrates a few possible distributions of normalized scores from the ensemble members in 2D. When ensemble members are ``good'', they assign higher scores to anomalies and push them to an extreme region of the score space as illustrated in case {\bf C1} in Figure~\ref{fig:idealized_rect}. This makes it easier to separate anomalies from nominals by a hyperplane. Most theoretical research on active learning for \emph{classification} \cite{kearns:1998,balcan:2007,kalai:2008,dasgupta:2009,balcan:2015} makes simplifying assumptions such as uniform data distribution over a unit sphere and with homogeneous (i.e., passing through the origin) hyperplanes. However, for \emph{anomaly detection}, arguably, the idealized setup is closer to case {\bf C2} (Figure~\ref{fig:idealized_circ}), where \emph{non-homogeneous} decision boundaries are more important. We present empirical evidence (Section~\ref{sec:experiments}), which shows that scores from the state-of-the-art \textit{Isolation Forest} (IFOR) detector are distributed in a similar manner as case {\bf C3} (Figure~\ref{fig:idealized_ifor}). {\bf C3} and {\bf C2} are similar in theory (for active learning) because both involve searching for the optimum non-homogeneous decision boundary. In all cases, the common theme is that when the ensemble members are ideal, then \emph{the scores of true anomalies tend to lie in the farthest possible location in the positive direction of the uniform weight vector ${\bf w}_{unif}$ \textbf{by design}}. Consequently, the average score for an instance across all ensemble members works well for anomaly detection. However, not all ensemble members are ideal in practice, and the true weight vector (${\bf w}^*$) is displaced by an angle $\theta$ from ${\bf w}_{unif}$. In large datasets, even a small misalignment between ${\bf w}_{unif}$ and ${\bf w}^*$ results in many false positives. While the performance of ensemble on the basis of the AUC metric may be high, the detector could still be impractical for use by analysts.

The property, that the misalignment is usually small, can be leveraged by active learning to learn the optimal weights efficiently. To understand this, observe that the top-ranked instances are close to the decision boundary and are therefore, in the uncertainty region. The key idea is to design a hyperplane that passes through the uncertainty region which then allows us to select query instances by uncertainty sampling. Selecting instances on which the model is uncertain for labeling is efficient for active learning \cite{cohn:1994,balcan:2007}. Specifically, greedily selecting instances with the highest scores is first of all more likely to reveal anomalies (i.e., true positives), and even if the selected instance is nominal (i.e., false positive), it still helps in learning the decision boundary efficiently. This is an important insight and has significant practical implications. {\bf Summary.} When detecting anomalies with an ensemble of detectors: (1) it is compelling to always apply active learning; (2) the greedy strategy of labeling top ranked instances is efficient, and is therefore a good yardstick for evaluating the performance of other querying strategies as well; and (3) learning a decision boundary with active learning that generalizes to unseen data helps in limited-memory or streaming data settings. The second point will be particularly significant when we evaluate a different querying strategy to enhance the diversity of discovered anomalies as part of this work.

\subsection{Tree-based Anomaly Detection Ensembles}
\label{sec:ensembles}

Tree-based ensemble detectors have several properties which make them ideal for active learning: (1) they can be employed to construct large ensembles inexpensively; (2) treating the nodes of tree as ensemble members allows us to both focus our feedback on fine-grained subspaces as well as increase the \emph{capacity} of the model; and (3) since some of the tree-based models such as \textit{Isolation Forest} (IFOR) \cite{liu:08}, \textit{HS Trees} (HST) \cite{tan:2011}, and \textit{RS Forest} (RSF) \cite{wu:2014} are state-of-the-art unsupervised detectors \cite{emmott:2015,domingues:2018}, it is a significant gain if their performance can be improved with minimal feedback \cite{das:2017}. We will focus mainly on IFOR in this work because it performed best across all datasets. However, we also present results on HST and RSF wherever applicable.

\textbf{IFOR} comprises of an ensemble of \textit{isolation} trees. Each leaf node is assigned a score proportional to the length of the path from the root to itself. By construction, leaves which correspond to subspaces that contain more anomalous instances have, on an average, smaller path lengths. In our implementation, we assign the leaf score as the {\em negative} path length; as a result, anomalous instances have higher scores than the nominals. We will represent the leaf-level scores by ${\bf d}$. After constructing the trees, we extract the {\bf leaf nodes as the ensemble members}. Each ensemble member assigns its leaf score as the anomaly score to an instance if the instance belongs to the corresponding subspace, else $0$. More details are included in the supplementary material.

\section{Active Learning Algorithms}
\label{sec:technical}

In this section, we first present a novel formalism called {\em compact description} that describes groups of instances compactly using a tree-based model. We then discuss a novel querying strategy that employs these descriptions to diversify the instances selected for labeling. When the selected instance(s) are labeled by an analyst, the model updates the weights of ensembles to be consistent with all the instances labeled so far. The algorithms to update the weights, in both batch and streaming data settings, are discussed next. In the \textit{batch} setting, the entire data is available at the outset; whereas, in the \textit{streaming} setting, the data comes as a continuous stream.

\subsection{Compact Description for a Group of Instances}
\label{sec:descriptions}
\begin{figure}[h]
	\centering
	\subfloat[Baseline]{
		\includegraphics[width=0.3\textwidth]{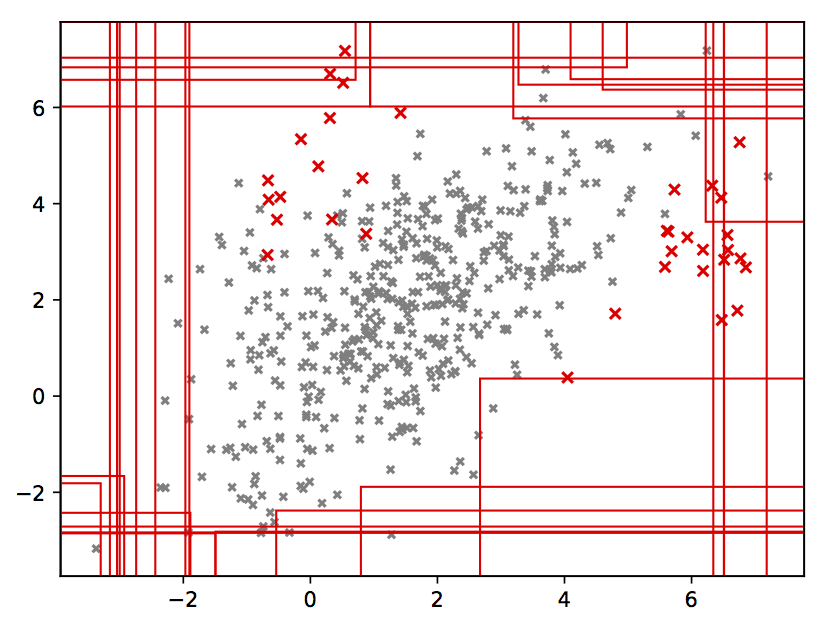}
		\label{fig:baseline_rects}}
	\subfloat[AAD]{
		\includegraphics[width=0.3\textwidth]{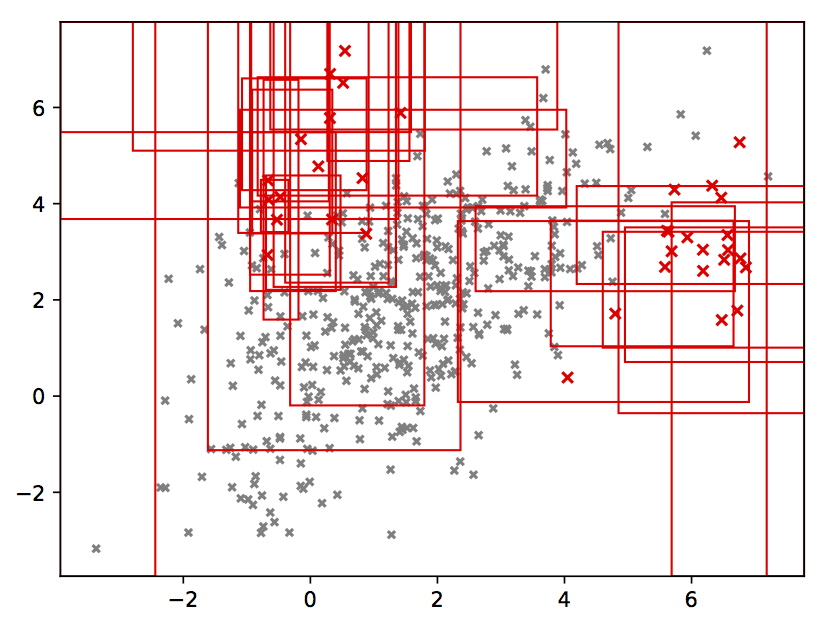}
		\label{fig:aad_rects}}
	\subfloat[Descriptions]{
		\includegraphics[width=0.3\textwidth]{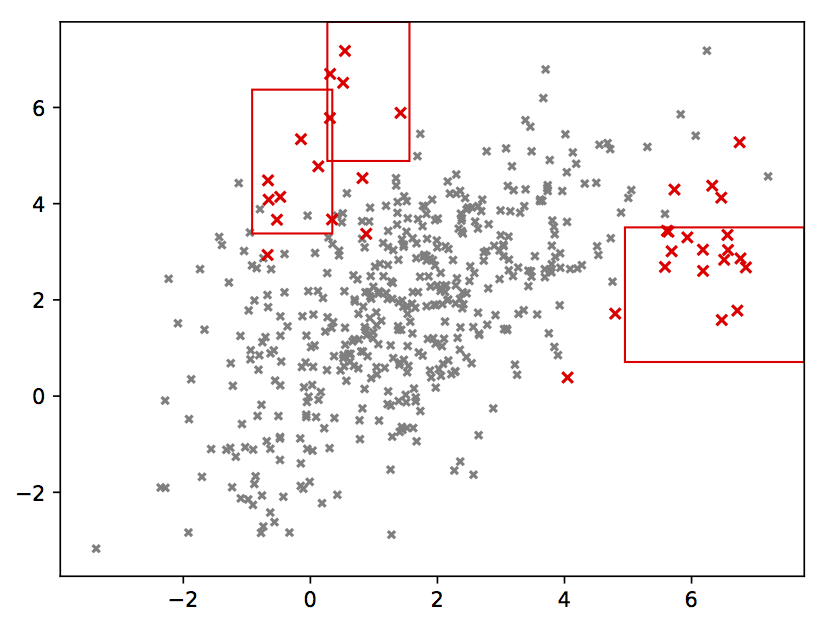}
		\label{fig:compact_rects}} \\[-1ex]
	\caption{Top $30$ subspaces ranked by ${\bf w}\circ{\bf d}$. Red points are anomalies. {\bf (a)} shows the top $30$ most relevant subspaces (w.r.t their \textit{anomalousness}) without any feedback. We can see that initially, these simply correspond to the exterior regions of the dataset. AAD \textbf{learns the true relevance of subspaces} via label feedback. {\bf (b)} shows that after incorporating the labels of $35$ instances, the subspaces around the labeled anomalies have emerged as the most relevant. {\bf (c)} shows the set of \textbf{important} subspaces which compactly cover all labeled anomalies. These were computed by solving Equation~\ref{eqn:opt}. Note that the compact subspaces only cover anomalies that were discovered in the $35$ feedback iterations. Anomalies which were not detected are likely to fall outside these compact subspaces.} \label{fig:rects}
\end{figure}

The tree-based model assigns a weight and an anomaly score to each leaf. We denote the vector of leaf-level anomaly scores by ${\bf d}$, and the overall anomaly scores of the subspaces (corresponding to the leaf-nodes) by ${\bf a} = \left[a_1, ..., a_m\right] = {\bf w}\circ{\bf d}$, where $\circ$ denotes element-wise product operation. The score $a_i$ provides a good measure of the \emph{relevance} of the $i$-th subspace. This relevance for each subspace is determined automatically through the label feedback. Our goal is to select a small subset of the most relevant and ``compact'' (by volume) subspaces which together contain all the instances in a group that we want to describe. We treat this problem as a specific instance of the \emph{set covering} problem. We illustrate this idea on a synthetic dataset in Figure~\ref{fig:rects}. This approach might be interpreted as a form of non-parametric clustering.

Let $\mathcal{Z}$ be the set of instances that we want to describe, where $|\mathcal{Z}|=p$. Let ${\bf s}_i$ be the $\delta$ most relevant subspaces (i.e., leaf nodes) which contain ${\bf z}_i \in \mathcal{Z}, i = {1, ..., p}$. Let $\mathcal{S}=s_1 \cup ... \cup s_p$ and $|\mathcal{S}|=k$. Denote the \textit{volumes} of the subspaces in $\mathcal{S}$ by the vector ${\bf v} \in \mathbb{R}^k$. Suppose ${\bf x} \in \{0, 1\}^k$ is a binary vector which contains $1$ in locations corresponding to the subspaces in $\mathcal{S}$ which are included in the covering set, and $0$ otherwise. Let ${\bf u}_i \in \{0, 1\}^k$ denote a vector for each instance ${\bf z}_i \in \mathcal{Z}$ which contains $1$ in all locations corresponding to subspaces in $s_i$. Let ${\bf U} = [{\bf u}_1^T, ..., {\bf u}_n^T]^T$. A compact set of subspaces ${\bf S}^*$ which contains (i.e., describes) all the candidate instances can be computed using the optimization formulation in Equation~\ref{eqn:opt}. We employ an off-the-shelf ILP solver (CVX-OPT) to solve this problem.
\begin{align}
{\bf S}^* &= \argmin_{{\bf x} \in \{0, 1\}^k} {\bf x} \cdot {\bf v} \label{eqn:opt} \\
\text{s.t. \;} & {\bf U} \cdot {\bf x} \geq {\bf 1} \text{ (where ${\bf 1}$ is a column vector of $p$ 1's)} \nonumber
\end{align}\
{\bf Applications of Descriptions.} Compact descriptions have multiple uses including: (1) Discovery of diverse classes of anomalies very quickly by querying instances from different subspaces of the description; and (2) Improved interpretability and explainability of anomalous instances. We assume that in a practical setting, the analyst(s) will be presented with instances along with their corresponding description(s). Additional information can be derived from the descriptions and shown to the analyst (e.g., number of instances in each description), which can help prioritize the analysis. In this work, we present empirical results on improving query diversity because it is easier to evaluate objectively than explanations or interpretability.

\subsection{Diversity-based Query Strategy}
\label{sec:query}

In Section~\ref{sec:insights}, we reasoned that the greedy strategy of selecting top-scored instances (referred as \texttt{Select-Top}) for labeling is efficient. However, this strategy might lack diversity in the types of instances presented to the analyst. It is likely that different types of instances belong to different subspaces in the original feature space. Our proposed strategy (\texttt{Select-Diverse}), which is described next, is intended to increase the diversity by employing tree-based ensembles to select groups of instances from subspaces that have minimum overlap.

\begin{figure}[h]
	\centering
	\subfloat[Regions]{\includegraphics[width=0.3\textwidth]{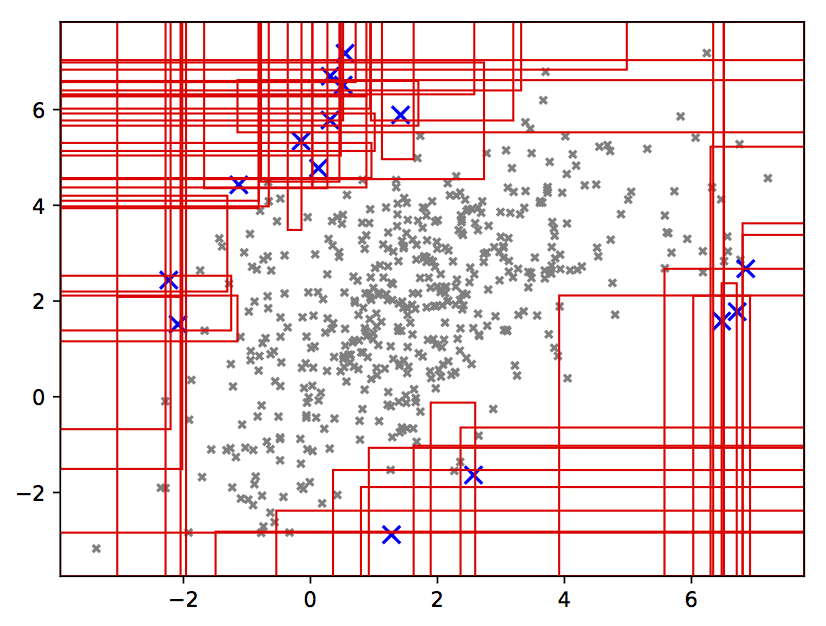}%
		\label{fig:query_regions}}
	\subfloat[Top]{\includegraphics[width=0.3\textwidth]{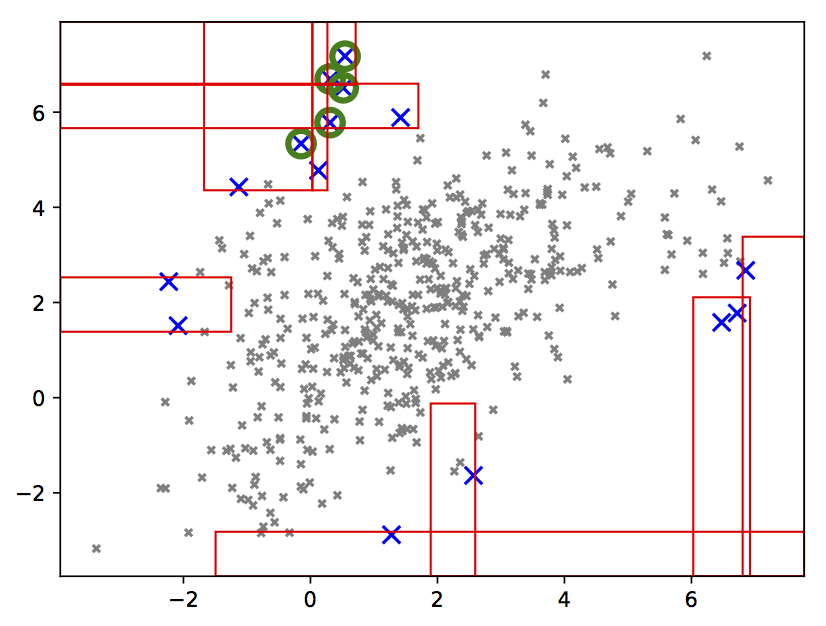}%
		\label{fig:query_baseline}}
	\subfloat[Diverse]{\includegraphics[width=0.3\textwidth]{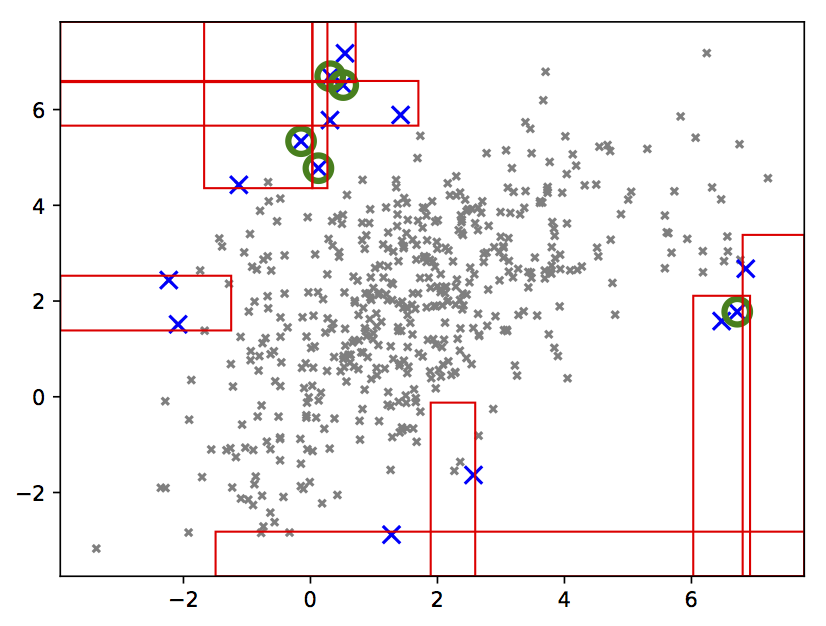}%
		\label{fig:query_diverse}} \\[-1ex]
	\caption{Illustration of compact description and diversity using IFOR. Most anomalous $15$ instances (blue checks) are selected as the query candidates. The red rectangles in {\bf (a)} form the union of the $\delta$ ($= 5$ works well in practice) most \emph{relevant} subspaces across each of the query candidates. {\bf (b)} and {\bf (c)} show the most ``compact'' set of subspaces which together cover all the query candidates. {\bf (b)} shows the most anomalous $5$ instances (green circles) selected by the greedy \texttt{Select-Top} strategy. {\bf (c)} shows the $5$ ``diverse'' instances (green circles) selected by \texttt{Select-Diverse}.}
	\label{fig:query}
\end{figure}

Assume that the analyst can label a batch of $b$ instances, where $b > 1$, in each feedback iteration. The following steps employ the compact descriptions to achieve this diversity (as illustrated in Figure~\ref{fig:query}): \textbf{1)} Select $\mathcal{Z}$ = a few top ranked instances as candidates (blue points in Figure~\ref{fig:query_regions}); \textbf{2)} Let ${\bf S}^*$ be the set of compact subspaces that contain $\mathcal{Z}$. (rectangles in Figures \ref{fig:query_baseline}) and \ref{fig:query_diverse}); \textbf{3)} Starting with the most anomalous instance in $\mathcal{Z}$, select one-by-one a set of $b$ instances denoted by ${\bf Q} \subset \mathcal{Z}$ for querying such that instances in ${\bf Q}$ belong to minimal overlapping regions (green circles in Figure~\ref{fig:query_diverse}).

\subsection{Algorithmic Framework to Update Weights}

\noindent {\bf Batch Active Learning (BAL).} We extend the AAD approach (based on LODA projections) \cite{das:2016} to update the weights for tree-based models. AAD assumes that (1) $\tau$ fraction of instances (i.e., $n\tau$) are anomalous, and (2) Anomalies should lie above the optimal hyperplane while nominals should lie below. AAD tries to satisfy these assumptions by enforcing constraints on the labeled examples while learning the weights of the hyperplane. If the anomalies are rare and we set $\tau$ to a small value, then the two assumptions make it more likely that the hyperplane will pass through the region of uncertainty. Our previous discussion then suggests that the optimal hyperplane can now be learned efficiently by greedily asking the analyst to label the most anomalous instance in each feedback iteration. We simplify the AAD formulation with a more scalable unconstrained optimization objective, and refer to this version as BAL. Crucially, the ensemble weights are updated with an intent to maintain the hyperplane in the region of uncertainty through the entire budget $B$. More details of the algorithm are included in the supplementary material.

\noindent {\bf Streaming Active Learning (SAL).} In the streaming case, we assume that the data is input to the algorithm continuously in \textit{windows} of size $K$ and is potentially unlimited. The framework for the streaming algorithm is straightforward and is included in the supplementary material. Initially, we train all the members of the ensemble with the first window of data. When a new window of data arrives, the underlying tree model is updated as follows: in case the model is an HST or RSF, only the node counts are updated while keeping the tree structures and weights unchanged; whereas, if the model is an IFOR, a subset of the current set of trees is replaced as shown in \texttt{Update-Model} (Algorithm~1). The updated model is then employed to determine which unlabeled instances to retain in memory, and which to ``forget''. This step, referred to as \texttt{Merge-and-Retain}, applies the simple strategy of retaining only the most anomalous instances among those in the memory and in the current window, and discards the rest. Next, the weights are fine-tuned with analyst feedback through an active learning loop similar to the batch setting with a small budget $Q$. Finally, the next window of data is read, and the process is repeated until the stream is empty or the total budget $B$ is exhausted. {\em In the rest of this section, we will assume that the underlying tree model is IFOR.}

When we replace a tree in \texttt{Update-Model}, its leaf nodes and corresponding weights get discarded. On the other hand, adding a new tree implies adding all its leaf nodes with weights initialized to a default value $v$. We first set $v = \frac{1}{\sqrt{m'}}$ where $m'$ is the total number of leaves in the new model, and then re-normalize the updated ${\bf w}$ to unit length.


SAL approach can be employed in two different situations: (1) \textbf{limited memory with no concept drift}, and (2) \textbf{streaming data with concept drift}. The type of situation determines how fast the model needs to be updated in \texttt{Update-Model}. If there is no concept drift, we need not update the model at all. If there is a large change in the distribution of data from one window to the next, then a large fraction of members need to be replaced. When we replace a member tree in our tree-based model, all its corresponding nodes along with their learned weights have to be discarded. Thus, some of the ``knowledge'' is lost with the model update. In general, it is hard to determine the true rate of drift in the data. One approach is to replace, in the \texttt{Update-Model} step, a reasonable number (e.g. 20\%) of older ensemble members with new members trained on new data. Although this ad hoc approach often works well in practice, a more principled approach is preferable.

\noindent {\bf Drift Detection Algorithm.} Algorithm~1 presents a principled methodology that employs KL-divergence (denoted by $D_{KL}$) to determine which trees should be replaced. The set of all leaf nodes in a tree are treated as a set of histogram bins which are then used to estimate the data distribution. We denote the total number of trees in the model by $T$, and the $t$-th tree by $\mathcal{T}_t$. When $\mathcal{T}_t$ is initially created with the first window of data, the same window is also used to initialize the {\em baseline} distribution for $\mathcal{T}_t$, denoted by ${\bf p}_t$ (\texttt{Get-Ensemble-Distribution} in Algorithm~3). After computing the baseline distributions for each tree, we estimate the $D_{KL}$ threshold $q_{KL}$ at the $\alpha_{KL}$ (typically $0.05$) significance level by sub-sampling (\texttt{Get-KL-Threshold} in Algorithm~2). When a new window is read, we first use it to compute the new distribution ${\bf q}_t$ (for $\mathcal{T}_t$). Next, if ${\bf q}_t$ differs {\em significantly} from ${\bf p}_t$ (i.e., $D_{KL}({\bf p}_t||{\bf q}_t) > q_{KL}$) for at least $2\alpha_{KL}T$ trees, then we replace all such trees with new ones created using the data from the new window. Finally, if any tree in the forest is replaced, then the baseline densities for all trees are recomputed with the data in the new window.

\begin{figure}[h]
	\centering
	\includegraphics[scale=0.8, angle=0, clip=true, trim=18mm 73mm 100mm 14mm]{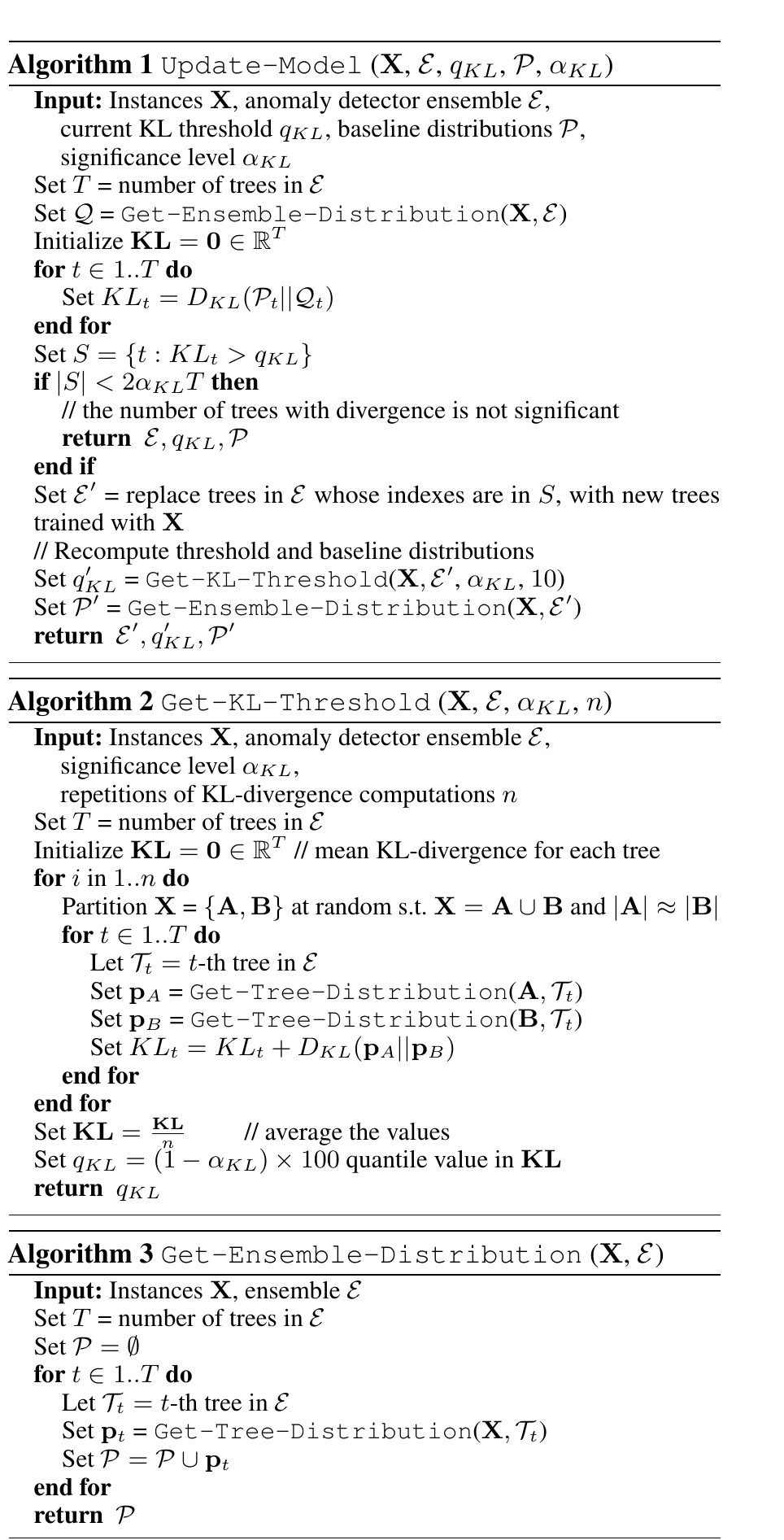}
\end{figure}

\section{Experiments and Results}
\label{sec:experiments}

\begin{figure}[h]
	\centering
	\includegraphics[scale=0.9, angle=0, clip=true, trim=15mm 119mm 35mm 100mm]{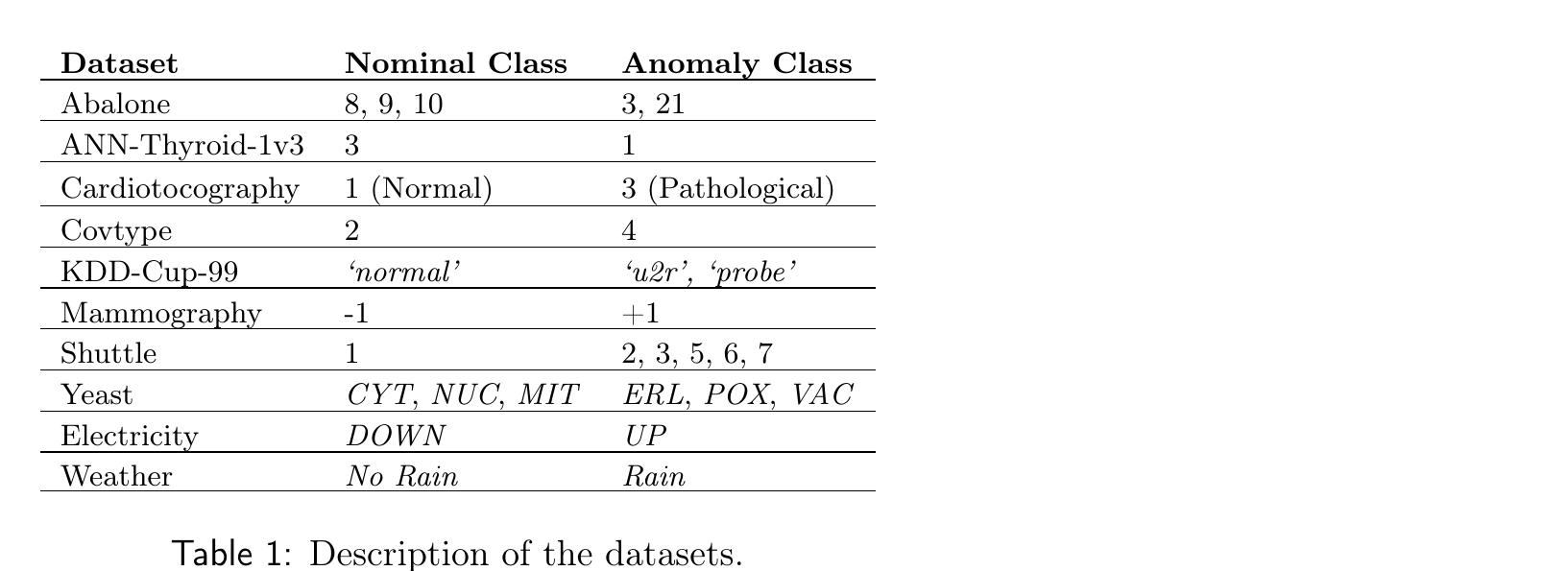}
\end{figure}

\begin{figure}[h]
	\centering
	\subfloat[ANN-Thyroid]{\includegraphics[width=0.33\textwidth]{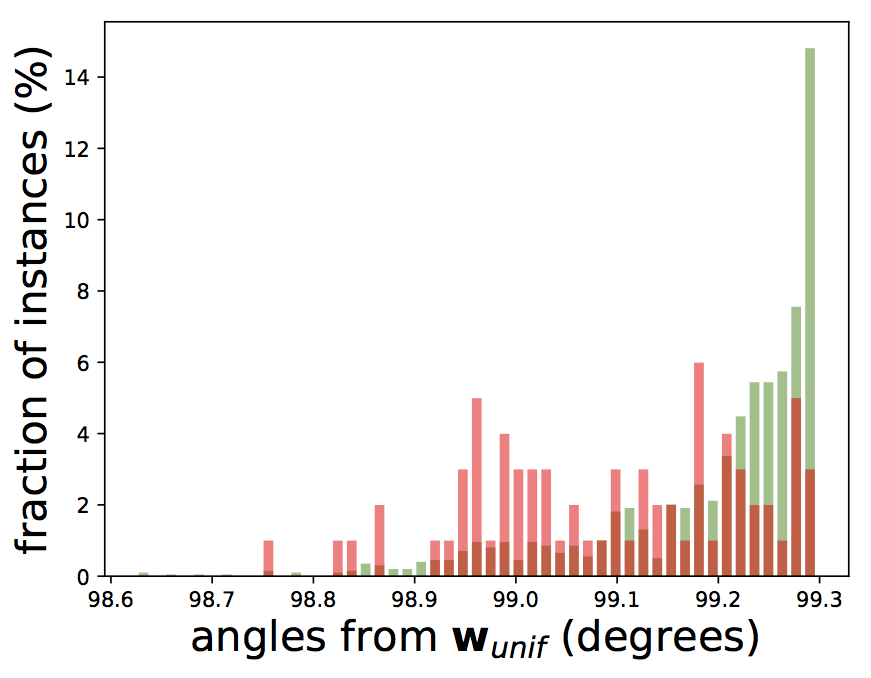}%
		\label{fig:angles_ann_thyroid_1v3}}
	\subfloat[Covtype]{\includegraphics[width=0.33\textwidth]{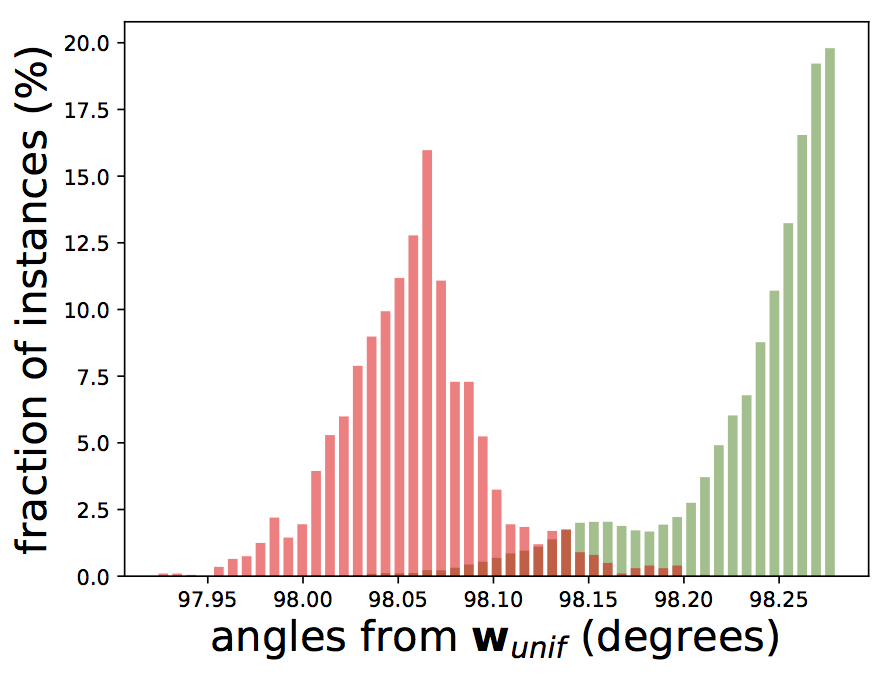}%
	\label{fig:angles_covtype}}
    \subfloat[Weather]{\includegraphics[width=0.33\textwidth]{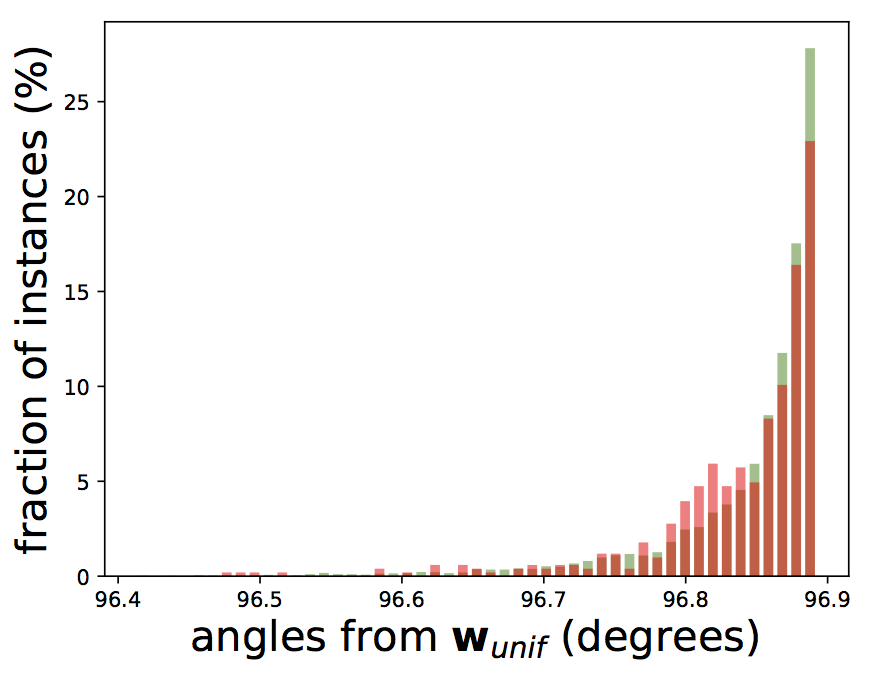}%
    	\label{fig:angles_weather}} \\[-1ex]
	\caption{Histogram distribution of the angles between score vectors from IFOR and ${\mathbf w}_{unif}$ for representative datasets (rest in Supplement). The red and green histograms show the angle distributions for anomalies and nominals respectively. Since the red histograms are closer to the left, anomalies are aligned closer to ${\mathbf w}_{unif}$.}
	\label{fig:angles}
\end{figure}

\noindent {\bf Experimental Setup.} All versions of BAL and SAL employ IFOR with the number of trees $T=100$ and subsample size $256$. The initial starting weights are denoted by ${\bf w}^{(0)}$. We normalize the score vector for each instance to unit length such that the score vectors lie on a unit sphere. This normalization helps adhere to the discussion in Section~\ref{sec:insights}, but is otherwise unnecessary. Figure~\ref{fig:angles} shows that ${\mathbf w}_{unif}$ tends to have a smaller angular separation from the normalized IFOR score vectors of anomalies than from those of nominals. This holds true for most of our datasets (Table~1). \textit{Weather} is a hard dataset for all anomaly detectors \cite{wu:2014}, as reflected in its angular distribution in Figure~\ref{fig:angles_weather}. In all our experiments, \texttt{Unsupervised Baseline} shows the number of anomalies detected without any feedback, i.e., using the uniform ensemble weights ${\mathbf w}_{unif}$; \texttt{BAL (No Prior - Unif)} and \texttt{BAL (No Prior - Rand)} impose no priors on the model, and start active learning with ${\mathbf w}^{(0)}$ set to ${\mathbf w}_{unif}$ and a random vector respectively; \texttt{BAL} sets ${\mathbf w}_{unif}$ as prior, and starts with ${\mathbf w}^{(0)}={\mathbf w}_{unif}$. For HST, we present two sets of results with batch input only: \texttt{HST-Batch} with original settings ($T=25$, depth=$15$, no feedback), and \texttt{HST-Batch (Feedback)} which supports feedback with BAL strategy (with $T=50$ and depth=$8$, a better setting for feedback). For RST, we present the results (\texttt{RST-Batch}) with only the original settings ($T=30$, depth=$15$) since it was not competitive with other methods on our datasets. We also compare the BAL variants with the AAD approach \cite{das:2016} in the batch setting (\texttt{AAD-Batch}). Batch data setting is the most optimistic for all algorithms. 

\noindent\textbf{Datasets and Evaluation Methodology.} We evaluate our algorithm on ten publicly available standard datasets (\cite{woods:1993},\cite{ditzler:2013},\cite{harries:1999}, UCI\cite{uci}) listed in Table~1 (details in supplementary materials). The anomaly classes in \textit{Electricity} and \textit{Weather} were down-sampled to be 5\% of the total. For each variant of the algorithm, we plot the percentage of the total number of anomalies shown to the analyst versus the number of instances queried; this is the most relevant metric for an analyst in any real-world application. Higher plot means the algorithm is better in terms of discovering anomalies. All results presented are averaged over 10 different runs and the error-bars represent $95\%$ confidence intervals.

\begin{figure*}[h]
	\centering
	\subfloat[Abalone]{\includegraphics[width=1.35in]{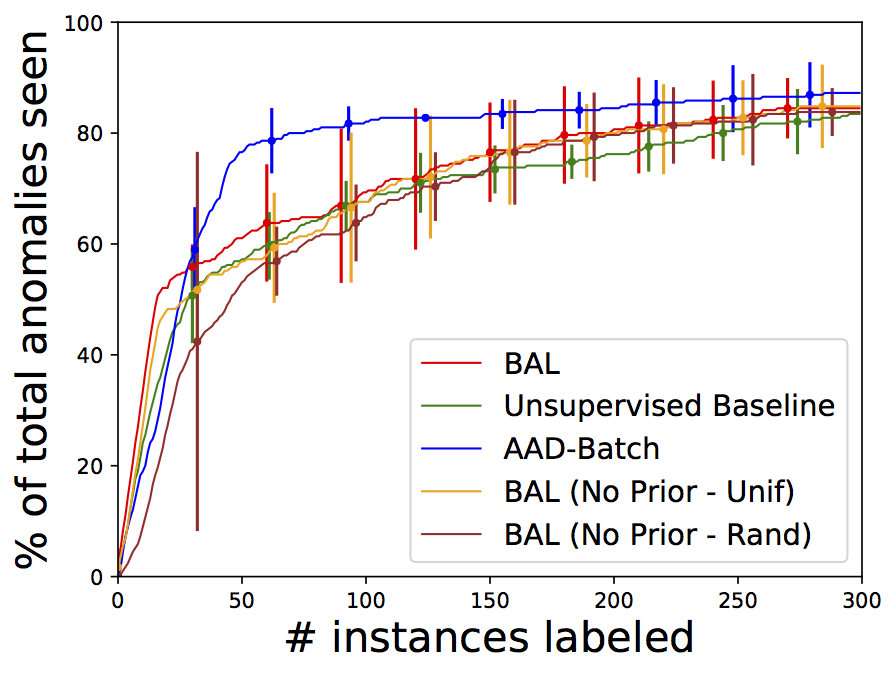}%
		\label{fig:batch_abalone}}
	\subfloat[ANN-Thyroid-1v3]{\includegraphics[width=1.35in]{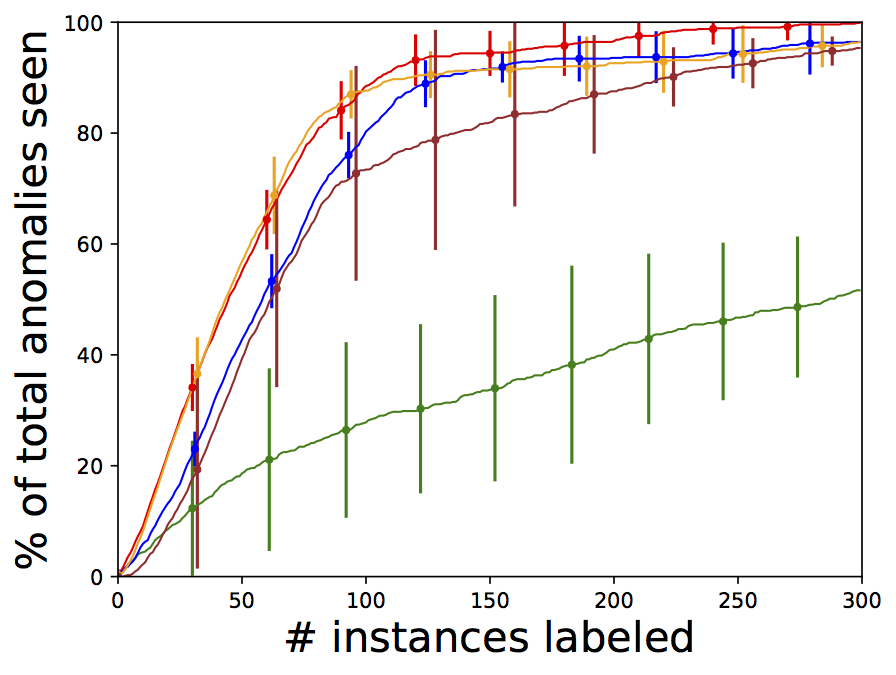}%
		\label{fig:batch_ann_thyroid_1v3}}
	\subfloat[Cardiotocography]{\includegraphics[width=1.35in]{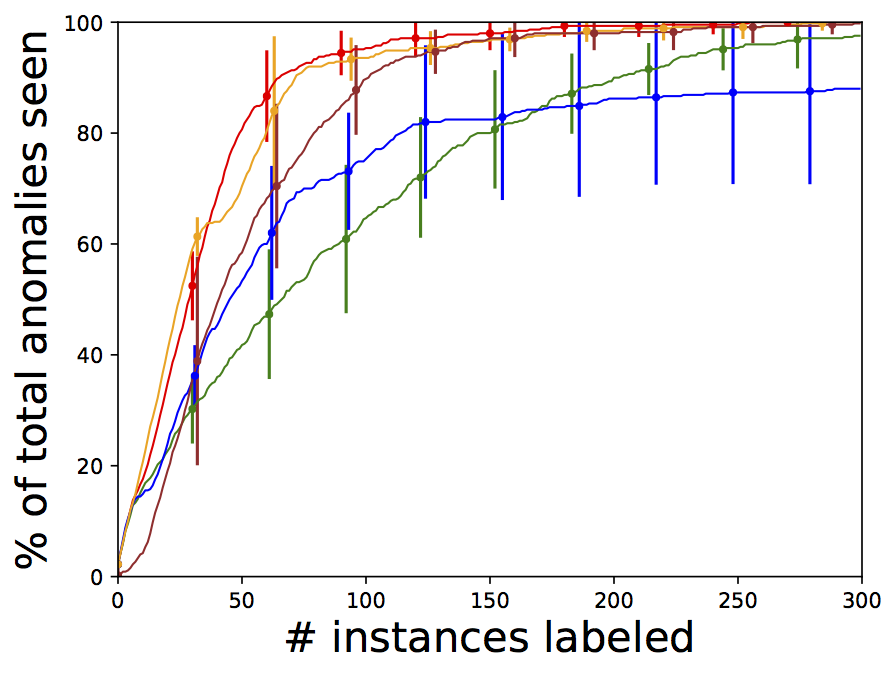}%
		\label{fig:batch_cardiotocography}}
	\subfloat[Yeast]{\includegraphics[width=1.35in]{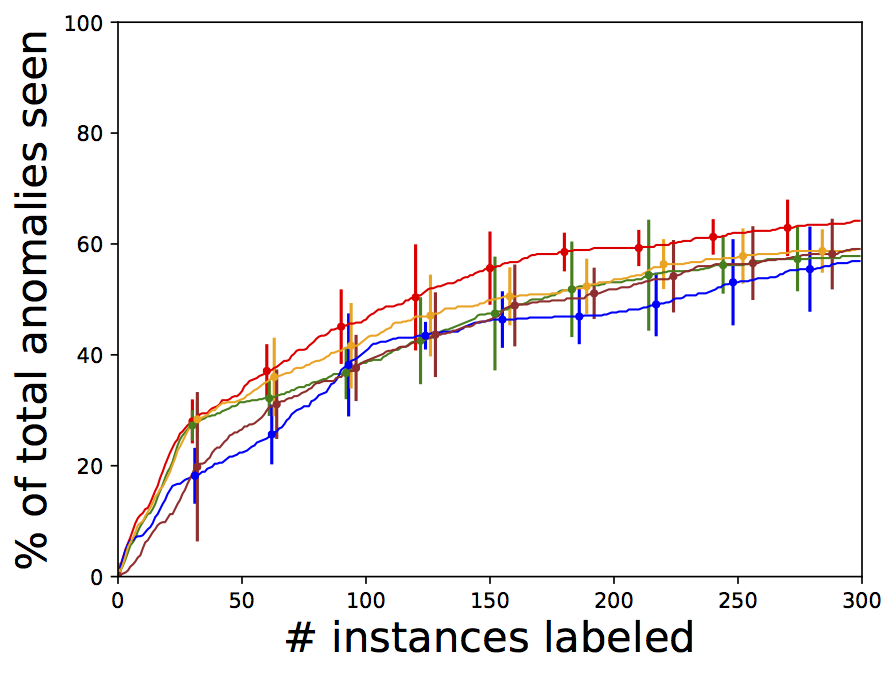}%
		\label{fig:batch_yeast}} \\[-1ex]
	\caption{Pct.~of total anomalies seen vs. the number of queries for the \textbf{smaller} datasets in the \textbf{batch} setting.}
	\label{fig:batch}
\end{figure*}
\begin{figure*}[h]
	\centering
	\subfloat[Covtype]{\includegraphics[width=1.35in]{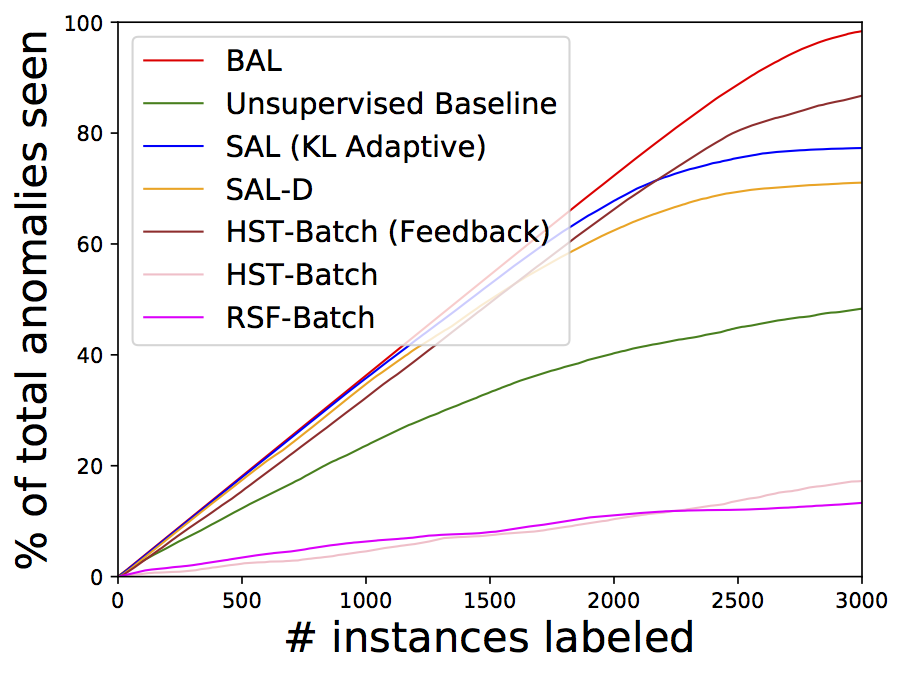}%
		\label{fig:stream_covtype}}
	\subfloat[Mammography]{\includegraphics[width=1.35in]{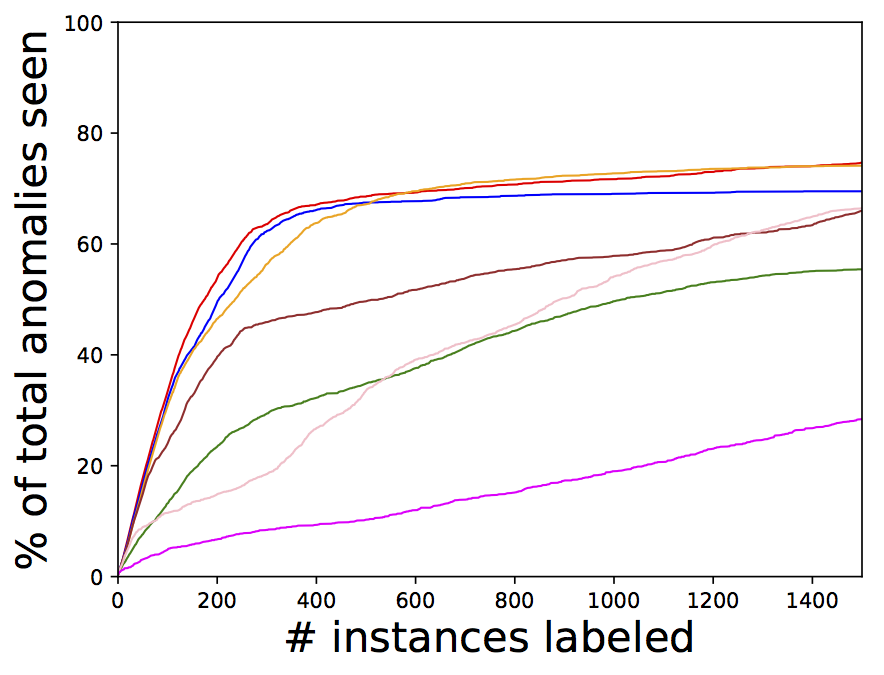}%
		\label{fig:stream_mammography}}
	\subfloat[KDD-Cup-99]{\includegraphics[width=1.35in]{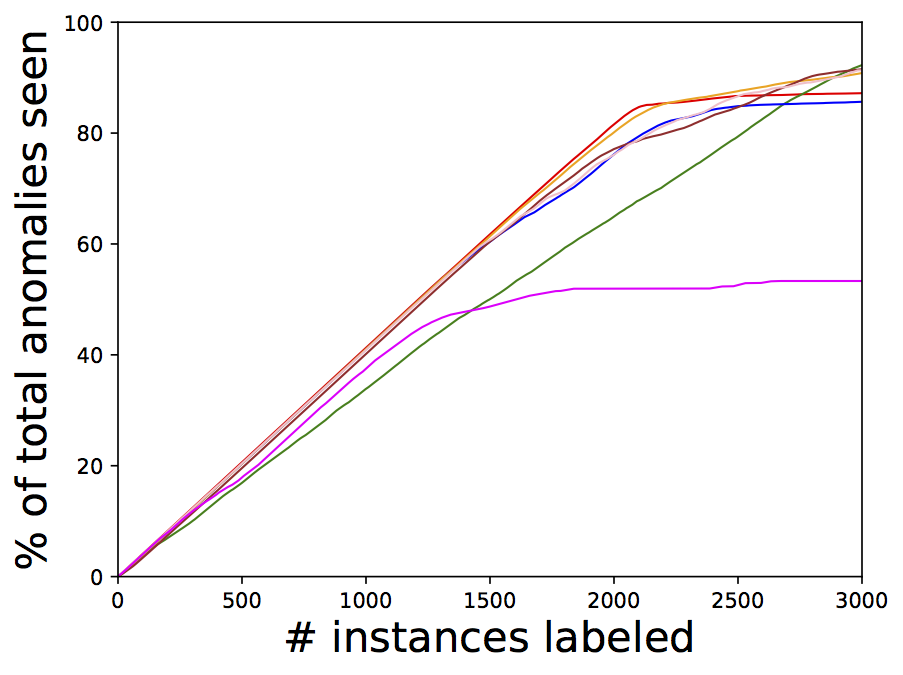}%
	\label{fig:stream_kddcup}}
	\subfloat[Shuttle]{\includegraphics[width=1.35in]{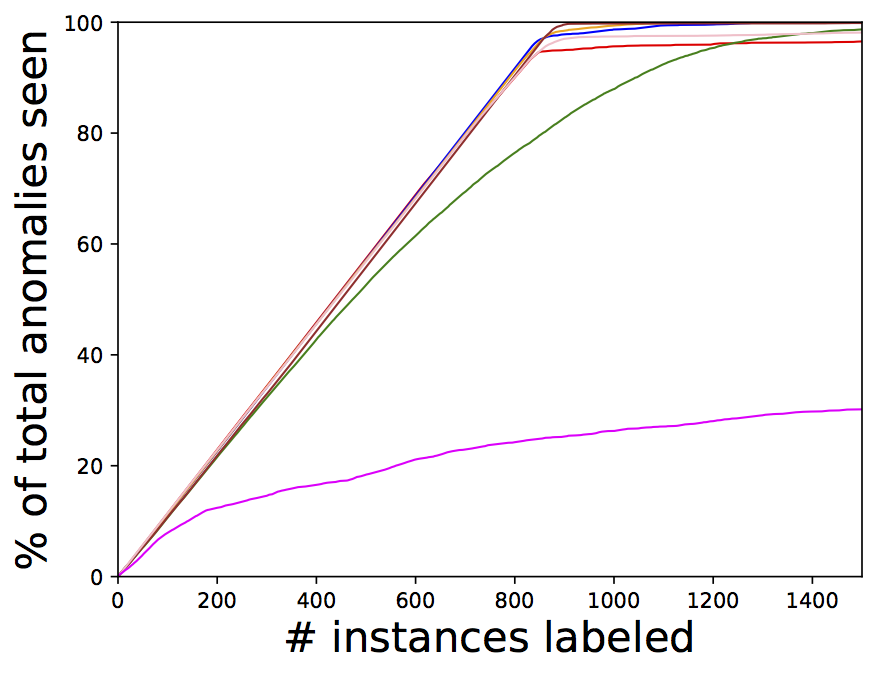}%
	\label{fig:stream_shuttle}} \\[-1ex]
	\caption{Pct.~of total anomalies seen vs. number of queries for the \textbf{larger datasets in the limited memory setting}. \texttt{SAL~(KL~Adaptive)} and \texttt{SAL-D} apply the \texttt{Select-Top} and \texttt{Select-Diverse} query strategies resp. \textit{Mammography}, \textit{KDD-Cup-99}, and \textit{Shuttle} have no significant drift. \textit{Covtype}, which has a higher drift, is included here for comparison because it is large.}
	\label{fig:limited_memory}
\end{figure*}

\subsection{Results for Diversified Query Strategy}
The diversity based query strategy \texttt{Select-Diverse} (Section~\ref{sec:query}) employs compact descriptions to select instances; 
therefore, the evaluation of its effectiveness is presented first. An interactive system can potentially ease the cognitive burden on the analysts by using descriptions to generate a ``summary'' of the anomalous instances.

\begin{figure}[h]
	\centering
	\subfloat[Class diversity]{\includegraphics[width=1.5in,height=1.2in]{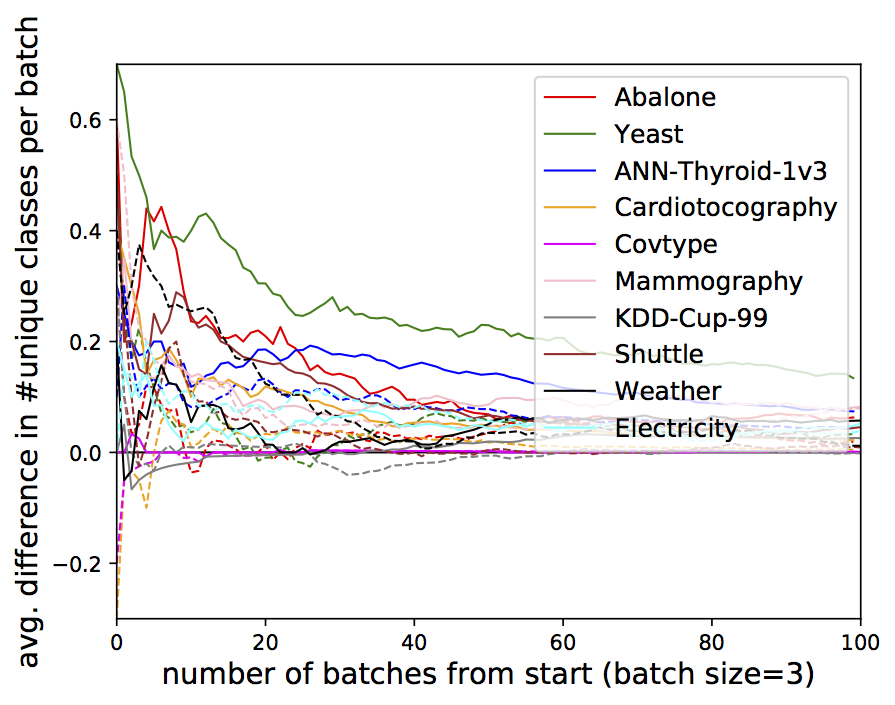}
		\label{fig:class_diversity}}
	\subfloat[\% anomalies seen]{\includegraphics[width=1.5in,height=1.2in]{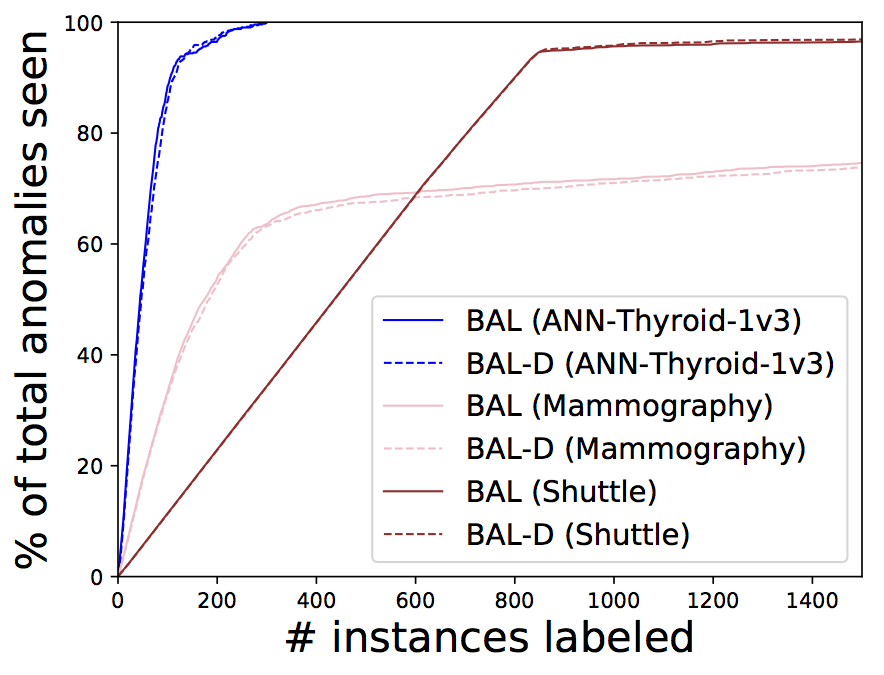}
		\label{fig:diversity_num_seen}} \\[-1ex]
	\caption{Query by diversity. The x-axis in {\bf (a)} shows the number of query batches (of batch size 3). The y-axis shows the difference in the number of unique classes seen averaged across all batches till a particular batch. The solid lines in {\bf (a)} show the average difference between unique classes seen with \texttt{BAL-D} and \texttt{BAL-T}; the dashed lines show the average difference between \texttt{BAL-D} and \texttt{BAL-R}.}
	\label{fig:query_diversity_all}
\end{figure}

We perform a post hoc analysis on the datasets with the knowledge of the original classes (Table~1). It is assumed that each class in a dataset represents a different data-generating process. To measure the diversity at any point in our feedback cycle, we compute the difference between the number of unique classes presented to the analyst per query batch averaged across all the past batches. The parameter $\delta$ for \texttt{Select-Diverse} was set to $5$ in all experiments. We compare three query strategies in the batch data setup: \texttt{BAL-T}, \texttt{BAL-D}, and \texttt{BAL-R}. \texttt{BAL-T} simply presents the top three most anomalous instances per query batch. \texttt{BAL-D} employs \texttt{Select-Diverse} to present three diverse instances out of the ten most anomalous instances. \texttt{BAL-R} presents three instances selected at random from the top ten anomalous instances. Finally, \texttt{BAL} greedily presents only the single most anomalous instance for labeling. We find that \texttt{BAL-D} presents a more diverse set of instances than both \texttt{BAL-T} (solid lines) as well as \texttt{BAL-R} (dashed lines) on most datasets. Figure~\ref{fig:diversity_num_seen} shows that the number of anomalies discovered (on representative datasets) with the diverse querying strategy is similar to the greedy strategy, i.e., no loss in anomaly discovery rate to improve diversity. The streaming data variants \texttt{SAL-*} have similar performance as the \texttt{BAL-*} variants.

\subsection{Results for Batch Active Learning (BAL)}
We set the budget $B$ to $300$ for all datasets in the batch setting. The results on the four smaller datasets \textit{Abalone, ANN-Thyroid-1v3, Cardiotocography,} and \textit{Yeast} are shown in Figure~\ref{fig:batch}. The performance is similar for the larger datasets. When the algorithm starts from sub-optimal initialization of the weights and with no prior knowledge (\texttt{BAL (No Prior - Rand)}), more number of queries are spent hunting for the first few anomalies, and thereafter detection improves significantly. When the weights are initialized to ${\mathbf w}_{unif}$, which is a reliable starting point (\texttt{BAL (No Prior - Unif)} and \texttt{BAL}), fewer queries are required to find the initial anomalies, and typically results in a lower variance in accuracy. Setting ${\mathbf w}_{unif}$ as prior in addition to informed initialization (\texttt{BAL}) performs better than without the prior (\texttt{BAL (No Prior - Unif)}) on \textit{Abalone, ANN-Thyroid-1v3}, and \textit{Yeast}. We believe this is because the prior helps guard against noise.

\subsection{Results for Streaming Active Learning (SAL)}
In all SAL experiments, we set the number of queries per window $Q=20$. The total budget $B$ and the stream window size $K$ for the datasets were set respectively as follows: \textit{Covtype} (3000, 4096), \textit{KDD-Cup-99} (3000, 4096), \textit{Mammography} (1500, 4096), \textit{Shuttle} (1500, 4096), \textit{Electricity} (1500, 1024), \textit{Weather} (1000, 1024). These values are reasonable w.r.t the dataset's size, the number of anomalies in it, and the rate of concept drift. The maximum number of unlabeled instances residing in memory is $K$. When the last window of data arrives, then active learning is continued with the final set of unlabeled data retained in the memory until the total budget $B$ is exhausted. The instances are streamed in the same order as they appear in the original public sources. When a new window of data arrives: \texttt{SAL~(KL~Adaptive)} dynamically determines which trees to replace based on KL-divergence, \texttt{SAL~(Replace $20\%$ Trees)} replaces $20\%$ oldest trees, and \texttt{SAL~(No~Tree~Replace)} creates the trees only once with the first window of data and only updates the weights of these fixed nodes with feedback thereafter.

\noindent\textbf{Limited memory setting with no concept drift: } The results on the four larger datasets are shown in Figure~\ref{fig:limited_memory}. The performance is similar to what is seen on the smaller datasets. Among the unsupervised algorithms in the batch setting, IFOR (\texttt{Unsupervised Baseline}) and HST (\texttt{HST-Batch}) are both competitive, and both are better than RSF (\texttt{RSF-Batch}). With feedback, \texttt{BAL} is consistently the best performer. HST with feedback (\texttt{HST-Batch (Feedback)}) always performs better than \texttt{HST-Batch}. The streaming algorithm with feedback, \texttt{SAL~(KL~Adaptive)}, significantly outperforms \texttt{Unsupervised Baseline} and is competitive with \texttt{BAL}. \texttt{SAL~(KL~Adaptive)} usually beats \texttt{HST-Batch (Feedback)} as well. \texttt{SAL-D} which presents a more diverse set of instances for labeling performs similar to \texttt{SAL~(KL~Adaptive)}. These results demonstrate that the feedback-tuned anomaly detectors generalize to unseen data.

\begin{figure}[t]
	\centering
	\subfloat[Covtype]{
		\includegraphics[width=0.32\textwidth]{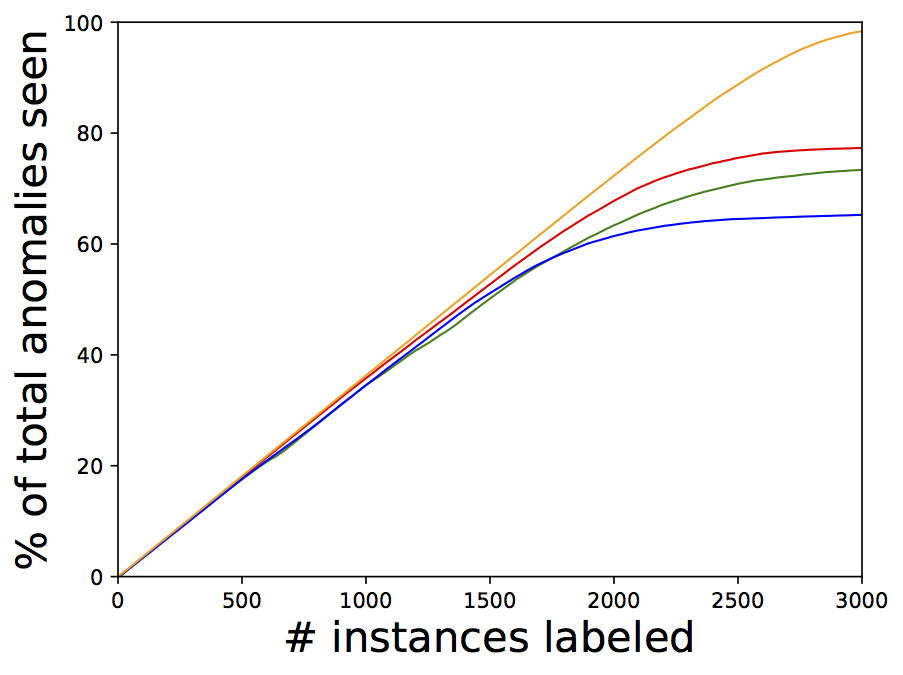}
		\label{fig:concept_drift_covtype_num}}
	\subfloat[Electricity]{
		\includegraphics[width=0.32\textwidth]{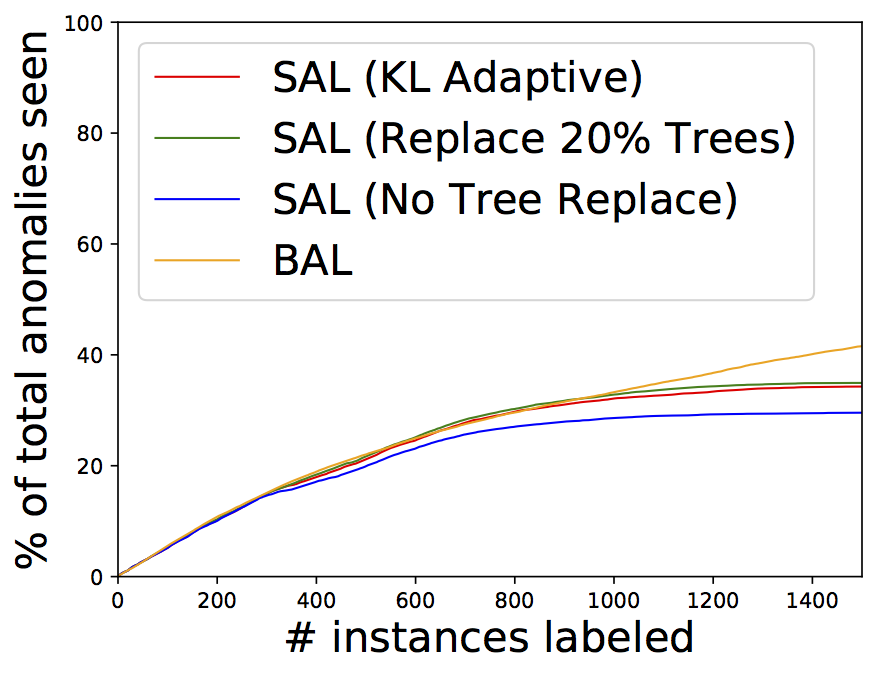}
		\label{fig:concept_drift_electricity_num}}
	\subfloat[Weather]{
		\includegraphics[width=0.32\textwidth]{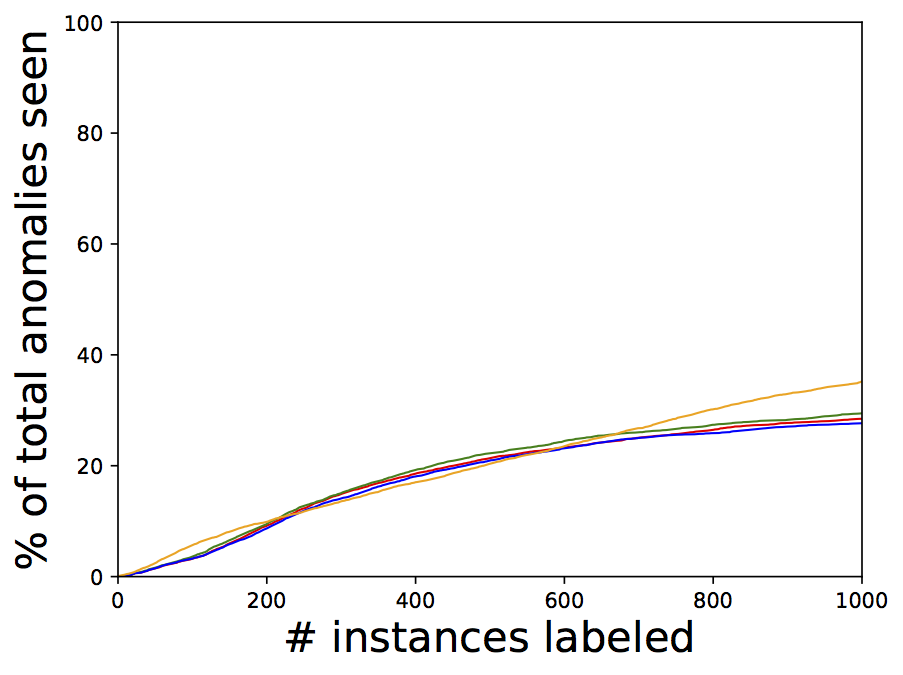}
		\label{fig:concept_drift_weather_num}} \\[-1ex]
	\caption{Integrated drift detection and label feedback with the Stream Active Learner (SAL) on the streaming datasets.}
	\label{fig:concept_drift_num}
\end{figure}

\begin{figure}[!h]
	\centering
	\subfloat[ANN-Thyroid-1v3]{
		\includegraphics[width=0.47\textwidth]{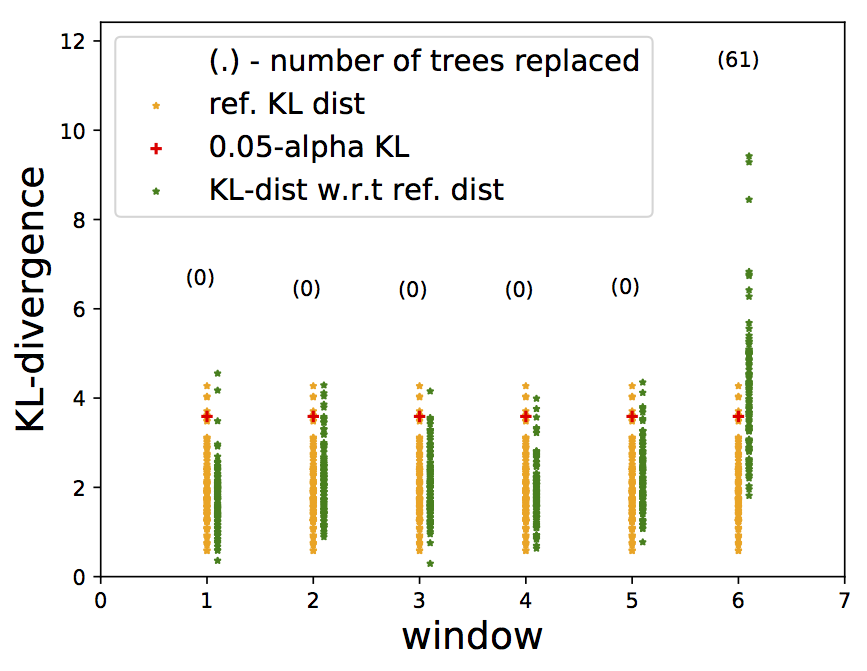}
		\label{fig:drift_ann_thyroid_1v3}}
	\subfloat[Covtype]{
		\includegraphics[width=0.47\textwidth]{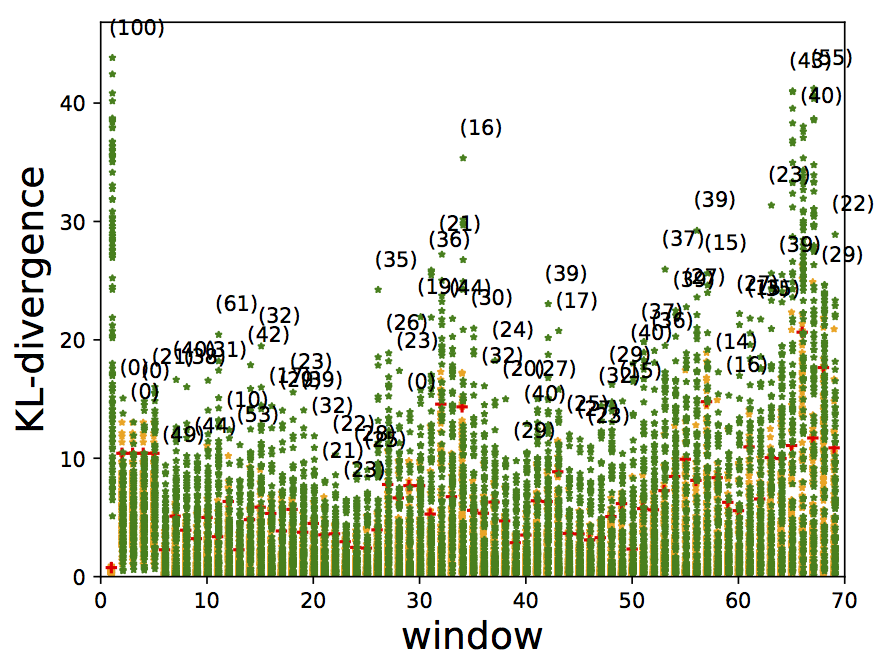}
		\label{fig:drift_covtype}}	\\[-1ex]
	\caption{Drift detection across windows. {\bf (a)} When there if no drift, such as in {\em ANN-Thyroid-1v3}, then no trees are replaced for most of the windows, and the older model is retained. {\bf (b)} If there is drift, such as in {\em Covtype}, then the trees are more likely to be replaced.}
	\label{fig:drift}
\end{figure}

\noindent\textbf{Streaming with Concept Drift:} The rate of drift determines how fast the model should be updated. If we update the model too fast, then we loose valuable knowledge gained from feedback which is still valid. On the other hand, if we update the model too slowly, then the model continues to focus on the stale subspaces based on past feedback. It is hard to find a common rate of update that works well across all datasets (such as replacing $20\%$ trees with each new window of data). Figure~\ref{fig:concept_drift_num} shows that the adaptive strategy (\texttt{SAL~(KL~Adaptive)}) which replaces obsolete trees using KL-divergence, as illustrated in Figure~\ref{fig:drift}, is robust and competitive with the best possible, i.e., \texttt{BAL}.

\section{Summary}
In this work, we called attention to an under-appreciated property of anomaly detector ensembles that makes them uniquely suitable for label-efficient active learning. We demonstrated the practical utility of this property through ensembles created from tree-based detectors. We also showed that the tree-based ensembles can be used to compactly describe groups of anomalous instances and diversify the queries to an analyst. Finally, we also developed a novel algorithm to detect the data drift in streaming setting and to take corrective actions. 

\bibliographystyle{aaai}
\bibliography{aaai2018}

\begin{thebibliography}{}

\bibitem[\protect\citeauthoryear{Aggarwal and Sathe}{2017}]{aggarwal:2017}
Aggarwal, C.~C., and Sathe, S.
\newblock 2017.
\newblock {\em Outlier Ensembles}.
\newblock Springer.

\bibitem[\protect\citeauthoryear{Balcan and Feldman}{2015}]{balcan:2015}
Balcan, M.~F., and Feldman, V.
\newblock 2015.
\newblock Statistical active learning algorithms for noise tolerance and
  differential privacy.
\newblock {\em Algorithmica} 72(1):282--315.

\bibitem[\protect\citeauthoryear{Balcan, Broder, and Zhang}{2007}]{balcan:2007}
Balcan, M.-F.; Broder, A.~Z.; and Zhang, T.
\newblock 2007.
\newblock Margin based active learning.
\newblock In {\em COLT}.

\bibitem[\protect\citeauthoryear{Breunig \bgroup et al\mbox.\egroup
  }{2000}]{breunig:00}
Breunig, M.~M.; Kriegel, H.-P.; Ng, R.~T.; and Sander, J.
\newblock 2000.
\newblock Lof: Identifying density-based local outliers.
\newblock In {\em ACM SIGMOD International Conference on Management of Data}.

\bibitem[\protect\citeauthoryear{Chandola, Banerjee, and
  Kumar}{2009}]{chandola:09}
Chandola, V.; Banerjee, A.; and Kumar, V.
\newblock 2009.
\newblock Anomaly detection: A survey.
\newblock {\em ACM Computing Surveys} 41(3):1–58.

\bibitem[\protect\citeauthoryear{Chen \bgroup et al\mbox.\egroup
  }{2017}]{chen:2017}
Chen, J.; Sathe, S.; Aggarwal, C.; and Turaga, D.
\newblock 2017.
\newblock Outlier detection with autoencoder ensembles.
\newblock In {\em SIAM International Conference on Data Mining}.

\bibitem[\protect\citeauthoryear{Cohn, Atlas, and Ladner}{1994}]{cohn:1994}
Cohn, D.; Atlas, L.; and Ladner, R.
\newblock 1994.
\newblock Improving generalization with active learning.
\newblock {\em Machine Learning} 15(2):201--221.

\bibitem[\protect\citeauthoryear{Das \bgroup et al\mbox.\egroup
  }{2016}]{das:2016}
Das, S.; Wong, W.-K.; Dietterich, T.~G.; Fern, A.; and Emmott, A.
\newblock 2016.
\newblock Incorporating expert feedback into active anomaly discovery.
\newblock In {\em IEEE ICDM}.

\bibitem[\protect\citeauthoryear{Das \bgroup et al\mbox.\egroup
  }{2017}]{das:2017}
Das, S.; Wong, W.-K.; Fern, A.; Dietterich, T.~G.; and Siddiqui, M.~A.
\newblock 2017.
\newblock Incorporating expert feedback into tree-based anomaly detection.
\newblock In {\em KDD IDEA Workshop}.

\bibitem[\protect\citeauthoryear{Dasgupta, Kalai, and
  Monteleoni}{2009}]{dasgupta:2009}
Dasgupta, S.; Kalai, A.~T.; and Monteleoni, C.
\newblock 2009.
\newblock Analysis of perceptron-based active learning.
\newblock {\em JMLR} 10:281--299.

\bibitem[\protect\citeauthoryear{Ditzler and Polikar}{2013}]{ditzler:2013}
Ditzler, G., and Polikar, R.
\newblock 2013.
\newblock Incremental learning of concept drift from streaming imbalanced data.
\newblock {\em IEEE Trans. on Knowl. and Data Eng.} 25(10):2283--2301.

\bibitem[\protect\citeauthoryear{Domingues \bgroup et al\mbox.\egroup
  }{2018}]{domingues:2018}
Domingues, R.; Filippone, M.; Michiardi, P.; and Zouaoui, J.
\newblock 2018.
\newblock A comparative evaluation of outlier detection algorithms: Experiments
  and analyses.
\newblock {\em Pattern Recognition} 74:406 -- 421.

\bibitem[\protect\citeauthoryear{Emmott \bgroup et al\mbox.\egroup
  }{2015}]{emmott:2015}
Emmott, A.; Das, S.; Dietterich, T.~G.; Fern, A.; and Wong, W.
\newblock 2015.
\newblock Systematic construction of anomaly detection benchmarks from real
  data.
\newblock {\em CoRR} abs/1503.01158.

\bibitem[\protect\citeauthoryear{Freund \bgroup et al\mbox.\egroup
  }{1997}]{freund:1997}
Freund, Y.; Seung, H.~S.; Shamir, E.; and Tishby, N.
\newblock 1997.
\newblock Selective sampling using the query by committee algorithm.
\newblock In {\em Machine Learning}.

\bibitem[\protect\citeauthoryear{G\"{o}rnitz \bgroup et al\mbox.\egroup
  }{2013}]{gornitz:2013}
G\"{o}rnitz, N.; Kloft, M.; Rieck, K.; and Brefeld, U.
\newblock 2013.
\newblock Toward supervised anomaly detection.
\newblock {\em Journal of Machine Learning Research} 46:235--262.

\bibitem[\protect\citeauthoryear{Guha \bgroup et al\mbox.\egroup
  }{2016}]{guha:2016}
Guha, S.; Mishra, N.; Roy, G.; and Schrijvers, O.
\newblock 2016.
\newblock Robust random cut forest based anomaly detection on streams.
\newblock In {\em ICML}.

\bibitem[\protect\citeauthoryear{Harries and of New
  South~Wales.}{1999}]{harries:1999}
Harries, M., and of~New South~Wales., U.
\newblock 1999.
\newblock {\em Splice-2 comparative evaluation [electronic resource] :
  electricity pricing}.
\newblock University of New South Wales, School of Computer Science and
  Engineering [Sydney].

\bibitem[\protect\citeauthoryear{He and Carbonell}{2008}]{he:2008}
He, J., and Carbonell, J.~G.
\newblock 2008.
\newblock Nearest-neighbor-based active learning for rare category detection.
\newblock In {\em NIPS}.

\bibitem[\protect\citeauthoryear{Kalai \bgroup et al\mbox.\egroup
  }{2008}]{kalai:2008}
Kalai, A.~T.; Klivans, A.~R.; Mansour, Y.; and Servedio, R.~A.
\newblock 2008.
\newblock Agnostically learning halfspaces.
\newblock {\em SIAM Journal on Computing} 37(6):1777--1805.

\bibitem[\protect\citeauthoryear{Kearns}{1998}]{kearns:1998}
Kearns, M.
\newblock 1998.
\newblock Efficient noise-tolerant learning from statistical queries.
\newblock {\em J. ACM} 45(6):983--1006.

\bibitem[\protect\citeauthoryear{Lazarevic and Kumar}{2005}]{lazarevic:05}
Lazarevic, A., and Kumar, V.
\newblock 2005.
\newblock Feature bagging for outlier detection.
\newblock In {\em KDD}.

\bibitem[\protect\citeauthoryear{Liu, Ting, and Zhou}{2008}]{liu:08}
Liu, F.~T.; Ting, K.~M.; and Zhou, Z.-H.
\newblock 2008.
\newblock Isolation forest.
\newblock In {\em IEEE ICDM}.

\bibitem[\protect\citeauthoryear{Monteleoni}{2006}]{monteleoni:2006}
Monteleoni, C.
\newblock 2006.
\newblock Efficient algorithms for general active learning.
\newblock In {\em COLT}.

\bibitem[\protect\citeauthoryear{Nissim \bgroup et al\mbox.\egroup
  }{2014}]{nissim:14}
Nissim, N.; Cohen, A.; Moskovitch, R.; Shabtai, A.; Edry, M.; Bar-Ad, O.; and
  Elovici, Y.
\newblock 2014.
\newblock Alpd: Active learning framework for enhancing the detection of
  malicious pdf files.
\newblock In {\em IEEE Joint Intelligence and Security Informatics Conference}.

\bibitem[\protect\citeauthoryear{Pevny}{2015}]{pevny:2015}
Pevny, T.
\newblock 2015.
\newblock Loda: Lightweight on-line detector of anomalies.
\newblock {\em Machine Learning} 102(2):275--304.

\bibitem[\protect\citeauthoryear{Rayana and Akoglu}{2016}]{rayana:2015}
Rayana, S., and Akoglu, L.
\newblock 2016.
\newblock Less is more: Building selective anomaly ensembles with application
  to event detection in temporal graphs.
\newblock In {\em SIAM International Conference on Data Mining}.

\bibitem[\protect\citeauthoryear{Senator \bgroup et al\mbox.\egroup
  }{2013}]{senator:2013}
Senator, T.~E.; Goldberg, H.~G.; Memory, A.; Young, W.~T.; Rees, B.; Pierce,
  R.; Huang, D.; Reardon, M.; Bader, D.~A.; Chow, E.; Essa, I.; Jones, J.;
  Bettadapura, V.; Chau, D.~H.; Green, O.; Kaya, O.; Zakrzewska, A.; Briscoe,
  E.; Mappus, R. I.~L.; McColl, R.; Weiss, L.; Dietterich, T.~G.; Fern, A.;
  Wong, W.-K.; Das, S.; Emmott, A.; Irvine, J.; Lee, J.-Y.; Koutra, D.;
  Faloutsos, C.; Corkill, D.; Friedland, L.; Gentzel, A.; and Jensen, D.
\newblock 2013.
\newblock Detecting insider threats in a real corporate database of computer
  usage activity.
\newblock In {\em KDD}.

\bibitem[\protect\citeauthoryear{Settles}{2010}]{settles:2010}
Settles, B.
\newblock 2010.
\newblock Active learning literature survey.
\newblock {\em University of Wisconsin, Madison} 52(55-66).

\bibitem[\protect\citeauthoryear{Stokes \bgroup et al\mbox.\egroup
  }{2008}]{stokes:2008}
Stokes, J.~W.; Platt, J.~C.; Kravis, J.; and Shilman, M.
\newblock 2008.
\newblock Aladin: Active learning of anomalies to detect intrusions.
\newblock {\em Technique Report. Microsoft Network Security Redmond, WA} 98052.

\bibitem[\protect\citeauthoryear{Tan, Ting, and Liu}{2011}]{tan:2011}
Tan, S.~C.; Ting, K.~M.; and Liu, T.~F.
\newblock 2011.
\newblock Fast anomaly detection for streaming data.
\newblock In {\em Proceedings of the Twenty-Second International Joint
  Conference on Artificial Intelligence - Volume Two},  1511--1516.

\bibitem[\protect\citeauthoryear{uci}{2007}]{uci}
2007.
\newblock {UC Irvine Machine Learning Repository}.
\newblock \url{http://archive.ics.uci.edu/ml/}.

\bibitem[\protect\citeauthoryear{Veeramachaneni \bgroup et al\mbox.\egroup
  }{2016}]{veeramachaneni:2016}
Veeramachaneni, K.; Arnaldo, I.; Cuesta-Infante, A.; Korrapati, V.; Bassias,
  C.; and Li, K.
\newblock 2016.
\newblock Ai2: Training a big data machine to defend.
\newblock {\em IEEE International Conference on Big Data Security}.

\bibitem[\protect\citeauthoryear{Woods \bgroup et al\mbox.\egroup
  }{1993}]{woods:1993}
Woods, K.~S.; Doss, C.~C.; Bowyer, K.~W.; Solka, J.~L.; Priebe, C.~E.; and
  Kegelmeyer, W.~P.
\newblock 1993.
\newblock Comparative evaluation of pattern recognition techniques for
  detection of microcalcifications in mammography.
\newblock {\em International Journal of Pattern Recognition and Artificial
  Intelligence} 07(06):1417--1436.

\bibitem[\protect\citeauthoryear{Wu \bgroup et al\mbox.\egroup
  }{2014}]{wu:2014}
Wu, K.; Zhang, K.; Fan, W.; Edwards, A.; and Philip, S.~Y.
\newblock 2014.
\newblock Rs-forest: A rapid density estimator for streaming anomaly detection.
\newblock In {\em Data Mining (ICDM), 2014 IEEE International Conference on},
  600--609.
\newblock IEEE.

\bibitem[\protect\citeauthoryear{Yan and Zhang}{2017}]{yan:2017}
Yan, S., and Zhang, C.
\newblock 2017.
\newblock Revisiting perceptron: Efficient and label-optimal learning of
  halfspaces.
\newblock In {\em NIPS}.

\end{thebibliography}

\section{Supplementary Materials}

\setcounter{algorithm}{3}
\setcounter{table}{1}

\subsection{Constructing Tree-based Ensembles}
\textbf{IFOR} comprises of an ensemble of \textit{isolation} trees. Each tree partitions the original feature space at random by recursively splitting an unlabeled dataset. At every tree-node, first a feature is selected at random, and then a split point for that feature is sampled uniformly at random. This partitioning operation is carried out until every instance reaches its own leaf node. The idea is that anomalous instances, which are generally isolated in the feature space, reach the leaf nodes faster by this partitioning strategy than nominal instances which belong to denser regions. Thus, the path from the root node is shorter to the leaves of anomalous instances than to the leaves of nominal instances. This path length is assigned as the unnormalized score for an instance by an isolation tree. After training an IFOR with $T$ trees, we extract the leaf nodes as the members of the ensemble. Such members could number in the thousands (typically $4000-7000$ when $T=100$). Assume that a leaf is at depth $l$ from the root. If an instance belongs to the partition defined by the leaf, it gets assigned a score $-l$ by the leaf, else $0$. As a result, anomalous instances receive higher scores on average than nominal instances. Since every instance belongs to only a few leaf nodes (equal to $T$), the score vectors are \textit{sparse} resulting in low memory and computation costs.

\begin{figure*}[!t]
	\centering
	\subfloat[IFOR]{\includegraphics[width=0.3\textwidth]{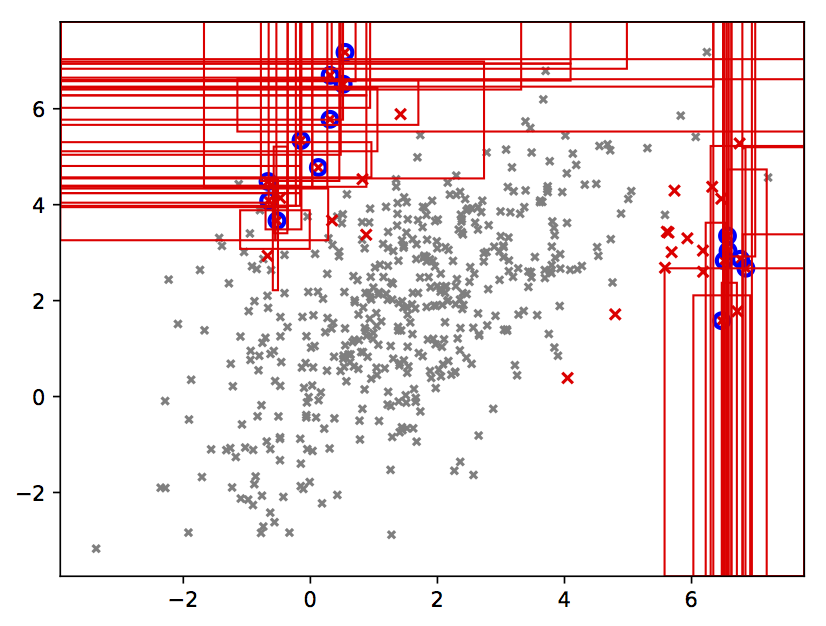}%
		\label{fig:iforest_regions}}
	\subfloat[HST (depth=$15$)]{\includegraphics[width=0.3\textwidth]{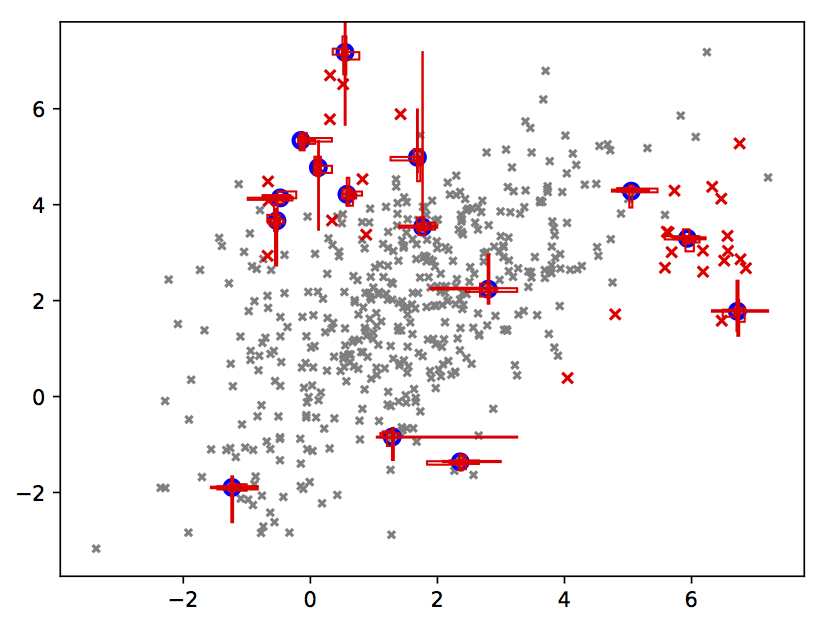}%
		\label{fig:hstrees_regions_15}}
	\subfloat[HST (depth=$8$)]{\includegraphics[width=0.3\textwidth]{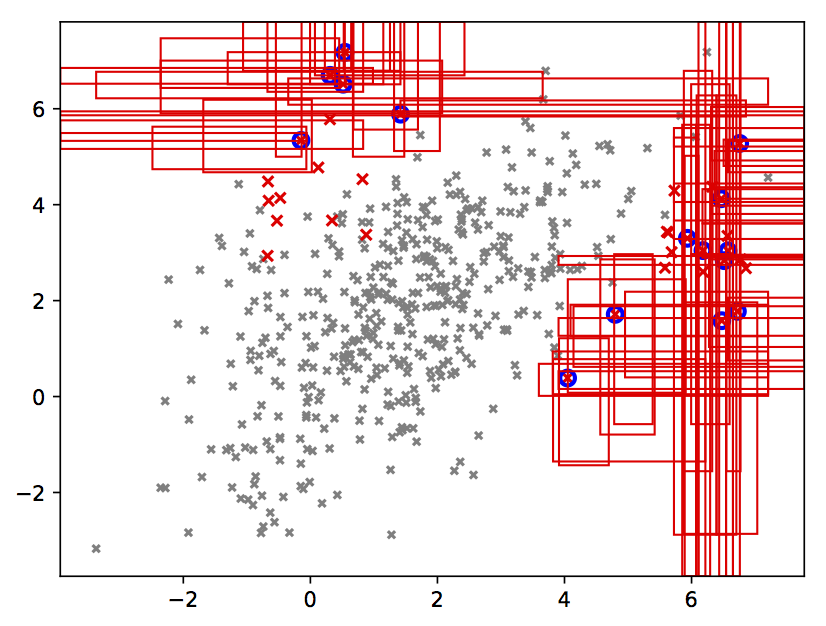}%
		\label{fig:hstrees_regions_8}}
	\caption{Illustration of differences among tree-based ensembles. The \textcolor{red}{red} rectangles show the union of the $5$ most anomalous subspaces across each of the $15$ most anomalous instances (\textcolor{blue}{blue}). These subspaces have the highest influence in propagating feedback across instances through gradient-based learning under our model. HST has fixed depth which needs to be high for accuracy (recommended $15$). IFOR has adaptive height and most anomalous subspaces are shallow. Higher depths are associated with smaller subspaces which are shared by fewer instances. As a result, feedback on any individual instance gets passed on to many other instances in IFOR, but to fewer instances in HST. RSF has similar behavior as HST. We set the depth for HST (and RSF) to $8$ (Figure~\ref{fig:hstrees_regions_8}) in our experiments in order to balance accuracy and feedback efficiency.}
	\label{fig:tree_differences}
\end{figure*}

\textbf{HST} and \textbf{RSF} apply different node splitting criteria than IFOR, and compute the anomaly scores on the basis of the sample counts and densities at the nodes. We apply log-transform to the leaf-level scores so that their unsupervised performance remains similar to the original and yet improves with feedback. The trees in HST and RSF have a fixed depth which needs to be larger in order to improve the accuracy. In contrast, IFOR has adaptive depth and most anomalous subspaces are shallow. Larger depths are associated with smaller subspaces which are shared by fewer instances. As a result, feedback on any individual instance gets passed on to very few instances in HST and RSF, but to many others in IFOR. Therefore, it is more efficient to incorporate feedback in IFOR than it is in HST or RSF (see Figure~\ref{fig:tree_differences}).

\subsection{Diversity-based Query Strategy}
\begin{figure*}[h]
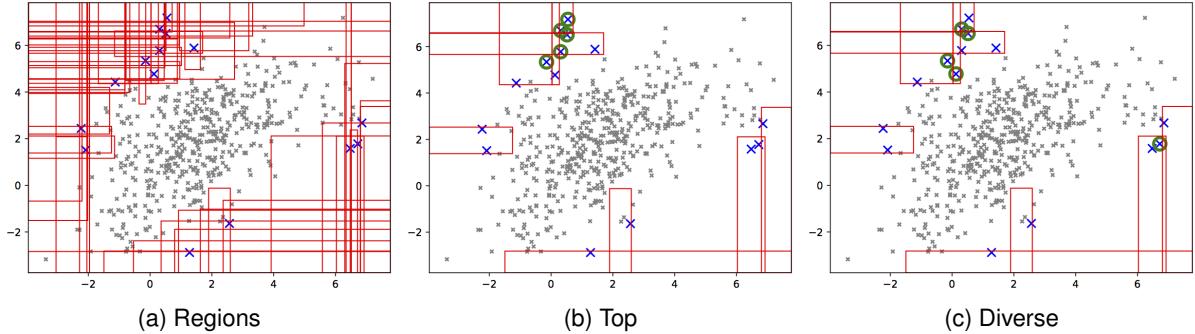

	\centering
	\subfloat[Regions]{\includegraphics[width=0.3\textwidth]{query_illustrate/query_candidate_regions_ntop5_100_trees.png}%
		\label{fig:query_regions_large}}
	\subfloat[Top]{\includegraphics[width=0.3\textwidth]{query_illustrate/query_compact_ntop5_100_trees_baseline.png}%
		\label{fig:query_baseline_large}}
	\subfloat[Diverse]{\includegraphics[width=0.3\textwidth]{query_illustrate/query_compact_ntop5_100_trees_aad.png}%
		\label{fig:query_diverse_large}} \\[-1ex]
	\caption{Illustration of compact description and diversity using IFOR. Most anomalous $15$ instances (\textcolor{blue}{blue} checks) are selected as the query candidates. The \textcolor{red}{red} rectangles in {\bf (a)} form the union of the $\delta$ ($= 5$ works well in practice) most \emph{relevant} subspaces across each of the query candidates. {\bf (b)} and {\bf (c)} show the most ``compact'' set of subspaces which together cover all the query candidates. {\bf (b)} shows the most anomalous $5$ instances (\textcolor{green}{green} circles) selected by the greedy \texttt{Select-Top} strategy. {\bf (c)} shows the $5$ ``diverse'' instances (\textcolor{green}{green} circles) selected by \texttt{Select-Diverse}.}
	\label{fig:query_large}
\end{figure*}

Figure~\ref{fig:query_large} illustrates the diversity-based query strategy. This is a larger size image of the same in the main paper and is repeated here for clarity.

\subsection{Active Learning Algorithms}
\label{app:active_algorithms}

\textbf{Batch Active Learning:} The batch active learning framework is presented in Algorithm~\ref{alg:batch}. BAL depends on only one hyper-parameter $\tau$.
We first define the hinge loss $\ell(q, {\mathbf w}; ({\mathbf z}_i, y_i))$ in Equation~\ref{eqn:aatploss} that penalizes the model when anomalies are assigned scores lower than $q$ and nominals higher. Equation~\ref{eqn:preflearn_aatp} then formulates the optimization problem for learning the optimal weights in Line~14 of the batch algorithm (Algorithm~1).

\begin{align}
& \ell(q, {\mathbf w}; ({\mathbf z}_i, y_i)) = \nonumber \\ 
& \left \{ 
\begin{array}{lr}
0 & {\mathbf w}\cdot {\mathbf z}_i \ge q \text{ and $y_i=+1$} \\
0 & {\mathbf w}\cdot {\mathbf z}_i < q \text{ and $y_i=-1$} \\
(q - {\mathbf w}\cdot {\mathbf z}_i) & {\mathbf w}\cdot {\mathbf z}_i < q \text{ and  $y_i=+1$} \\
({\mathbf w}\cdot {\mathbf z}_i - q) & {\mathbf w}\cdot {\mathbf z}_i \ge q \text{ and $y_i=-1$}
\end{array} 
\right. \label{eqn:aatploss}
\end{align}

\begin{align}
{\mathbf w}^{(t)} &= \argmin_{{\mathbf w}} \sum_{s \in \{-, +\}} \left( \frac{1}{|{\mathbf H}_s|} \sum_{{\mathbf z}_i \in {\mathbf H}_s}\ell(\hat{q}_{\tau}( {{\mathbf w}^{(t-1)}}), {\mathbf w}; ({\mathbf z}_i, y_i)) \right. \nonumber \\
& \left. \qquad \qquad + \frac{1}{|{\mathbf H}_s|} \sum_{{\mathbf z}_i \in {\mathbf H}_s}\ell({\mathbf z}_{\tau}^{(t-1)} \cdot {\mathbf w}, {\mathbf w}; ({\mathbf z}_i, y_i)) \right) \nonumber \\
& \qquad \qquad + \lambda^{(t)}\|{\mathbf w} - {\mathbf w}_{unif}\|^2 \label{eqn:preflearn_aatp} \\
& \text{where, } \mathbf{w}_{unif} = [\frac{1}{\sqrt{m}},\ldots,\frac{1}{\sqrt{m}}]^T, \text{ and,} \nonumber \\ 
&{\mathbf z}_{\tau}^{(t-1)} \text{ and } \hat{q}_{\tau}({{\mathbf w}^{(t-1)}}) \text{ are computed by ranking} \nonumber \\
&\text{anomaly scores with ${\mathbf w} = {\mathbf w}^{(t-1)}$} \nonumber
\end{align}

$\lambda^{(t)}$ determines the influence of the prior. For the batch setup, we set $\lambda^{(t)}=\frac{0.5}{|{\bf H}_+|+|{\bf H}_-|}$ such that the prior becomes less important as more instances are labeled. When there are no labeled instances, $\lambda^{(t)}$ is set to $\frac{1}{2}$. The third and fourth terms of Equation~\ref{eqn:preflearn_aatp} encourage the scores of anomalies in $\mathbf{H}_+$ to be higher than that of ${\mathbf z}_{\tau}^{(t-1)}$ (the $n\tau$-th ranked instance from the previous iteration), and the scores of nominals in $\mathbf{H}_-$ to be lower than that of ${\mathbf z}_{\tau}^{(t-1)}$. We employ gradient descent to learn the optimal weights ${\mathbf w}$ in Equation~\ref{eqn:preflearn_aatp}. Our prior knowledge that ${\bf w}_{unif}$ is a good prior provides a good initialization for gradient descent. We later show empirically that ${\bf w}_{unif}$ is a better starting point than random weights.

\begin{algorithm}
	\footnotesize
	\caption{\texttt{Batch-AL} ($B$, ${\mathbf w}^{(0)}$, ${\mathbf H}$, ${\mathbf H}_+$, ${\mathbf H}_-$)}
	\label{alg:batch}
	\begin{algorithmic}[14]
		\STATE \textbf{Input:} Budget $B$, initial weight ${\mathbf w}^{(0)}$, unlabeled instances ${\mathbf H}$, 
		\STATE \hspace{0.1in} labeled instances ${\mathbf H}_+$ and ${\mathbf H}_-$
		\STATE Set $t=0$
		\WHILE{$t \leq B$}
		\STATE Set $t = t + 1$
		\STATE Set ${\mathbf a} = {\mathbf H} \cdot {\mathbf w}$ (i.e., ${\mathbf a}$ is the vector of anomaly scores)
		\STATE Let ${\bf q} = {\mathbf z}_i$, where $i = \argmax_{i}(a_i)$
		\STATE Get $y_i \in \{-1,+1\}$ for ${\mathbf q}$ from analyst
		\IF{$y_i = +1$}
		\STATE Set ${\mathbf H}_+ = \{{\bf z}_i\} \cup {\mathbf H}_+$
		\ELSE
		\STATE Set ${\mathbf H}_- = \{{\bf z}_i\} \cup {\mathbf H}_-$
		\ENDIF
		\STATE Set ${\mathbf H} = {\mathbf H} \setminus {\mathbf z}_i$
		\STATE ${\mathbf w}^{(t)}$ = learn new weights; normalize $\|{\mathbf w}^{(t)}\|=1$ \label{alg:batch:weightupdate_}
		\ENDWHILE
		\RETURN ${\mathbf w}^{(t)}$, ${\mathbf H}$, ${\mathbf H}_+$, ${\mathbf H}_-$
	\end{algorithmic}
\end{algorithm}

\noindent \textbf{Stream Active Learning:} The stream active learning framework is presented in Algorithm~\ref{alg:stream}. In all the SAL experiments, we set the number of queries per window $Q=20$, and $\lambda^{(t)}=\frac{1}{2}$.
\label{app:stream_active_algorithms}
\begin{algorithm}[h!]
	\footnotesize
	\caption{\texttt{Stream-AL} ($K$, $B$, $Q$, $\mathcal{E}^{(0)}$, ${\mathbf X}_0$, ${\mathbf w}^{(0)}$, $\alpha_{KL}$)}
	\label{alg:stream}
	\begin{algorithmic}
		\STATE \textbf{Input:} Stream window size $K$, total budget $B$, 
		\STATE \hspace{0.1in} queries per window $Q$, anomaly detector ensemble $\mathcal{E}^{(0)}$, 
		\STATE \hspace{0.1in} initial instances ${\mathbf X}_0$ (used to create $\mathcal{E}^{(0)}$), initial weight ${\mathbf w}^{(0)}$,
		\STATE \hspace{0.1in} significance level $\alpha_{KL}$
		\STATE
		\STATE Set ${\mathbf H} = {\mathbf H}_+ = {\mathbf H}_- = \emptyset$
		\STATE // initialize KL-divergence baselines
		\STATE Set $q_{KL}^{(0)}$ = \texttt{Get-KL-Threshold}(${\mathbf X}_0$, $\mathcal{E}^{(0)}$, $\alpha_{KL}$, $10$)
		\STATE Set $\mathcal{P}^{(0)}$ = \texttt{Get-Ensemble-Distribution}(${\mathbf X}_0$, $\mathcal{E}^{(0)}$)
		\STATE
		\STATE Set $t=0$
		\WHILE{$<$stream is not empty$>$}
		\STATE Set $t = t + 1$
		\STATE Set ${\mathbf X}_t$ = $K$ instances from stream
		\STATE Set ${\mathbf H}_t$ = transform ${\mathbf X}_t$ to ensemble features
		\STATE 
		\STATE // \texttt{Update-Model} either updates node counts (e.g., for HST and RSF), 
		\STATE // or replaces a fraction of the oldest trees in $\mathcal{E}$ with new
		\STATE // ones constructed with ${\mathbf X}_t$ (e.g., IFOR)
		\STATE Set $\mathcal{E}^{(t)}$, $q_{KL}^{(t)}$, $\mathcal{P}^{(t)}$ = \texttt{Update-Model}(${\mathbf X}_t$, $\mathcal{E}^{(t-1)}$, $q_{KL}^{(t-1)}$, $\mathcal{P}^{(t-1)}$, $\alpha_{KL}$)
		\STATE 
		\STATE // \texttt{Merge-and-Retain}(${\mathbf w}$, ${\mathbf H}$, $K$) retains $K$ 
		\STATE // most anomalous instances in ${\mathbf H}$
		\STATE Set ${\mathbf H} =$ \texttt{Merge-and-Retain}(${\mathbf w}^{(t-1)}$, $\{{\mathbf H} \cup {\mathbf H}_t\}$, $K$)
		\STATE Set ${\mathbf w}^{(t)}$, ${\mathbf H}$, ${\mathbf H}_+$, ${\mathbf H}_-$ = \texttt{Batch-AL}($Q$, ${\mathbf w}^{(t-1)}$, ${\mathbf H}$, ${\mathbf H}_+$, ${\mathbf H}_-$)
		\ENDWHILE
	\end{algorithmic}
\end{algorithm}

\begin{algorithm}[h!]
	\footnotesize
	\caption{\texttt{Get-Tree-Distribution} (${\mathbf X}$, $\mathcal{T}$)}
	\label{alg:get_tree_distribution}
	\begin{algorithmic}
		\STATE \textbf{Input:} Instances ${\mathbf X}$, tree $\mathcal{T}$
		\STATE Set ${\mathbf p} = $ distribution of instances in ${\mathbf X}$ at the leaves of $\mathcal{T}$
		\RETURN ${\mathbf p}$
	\end{algorithmic}
\end{algorithm}

The KL-divergence approach applies only to trees whose structures are dependent on the input sample distributions (such as IFOR), and not to those whose structures are only a function of feature ranges (such as HST, RSF). In IFOR, the leaf depths are indicative of the density, and therefore, their anomalousness. If the isolation trees are not appropriately updated, then the true anomalousness of the subspaces will not be reflected in their structure.

\subsection{Histogram of Angle Distributions for IFOR}
\begin{figure*}[h]
	\centering
	\subfloat[Abalone]{\includegraphics[width=0.24\textwidth]{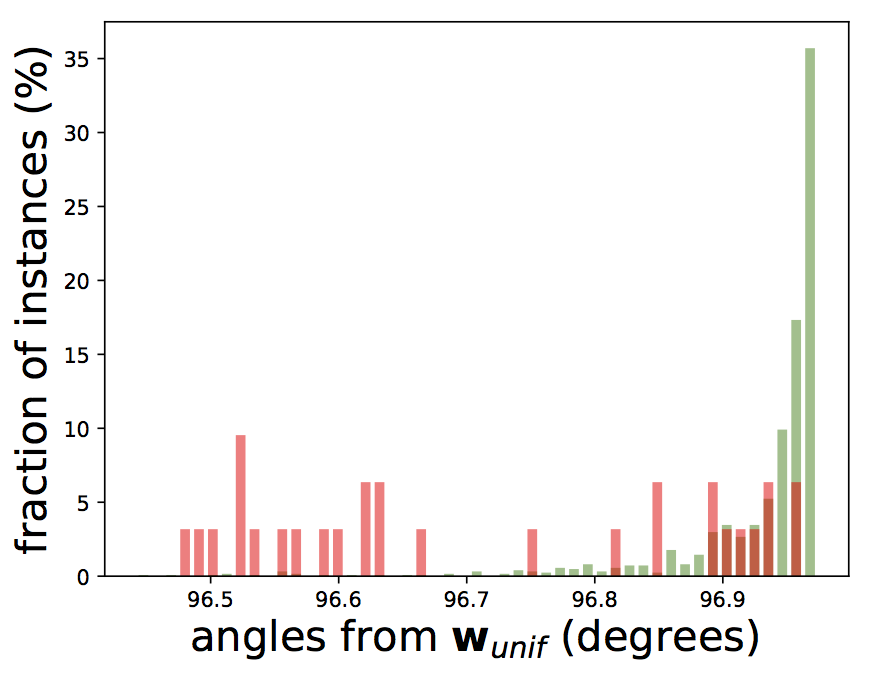}%
	\label{fig:angles_abalone}}
	\subfloat[ANN-Thyroid]{\includegraphics[width=0.24\textwidth]{angles/angles_ann_thyroid_1v3_iforest.png}%
		\label{fig:angles_ann_thyroid_1v3}}
	\subfloat[Cardiotocography]{\includegraphics[width=0.24\textwidth]{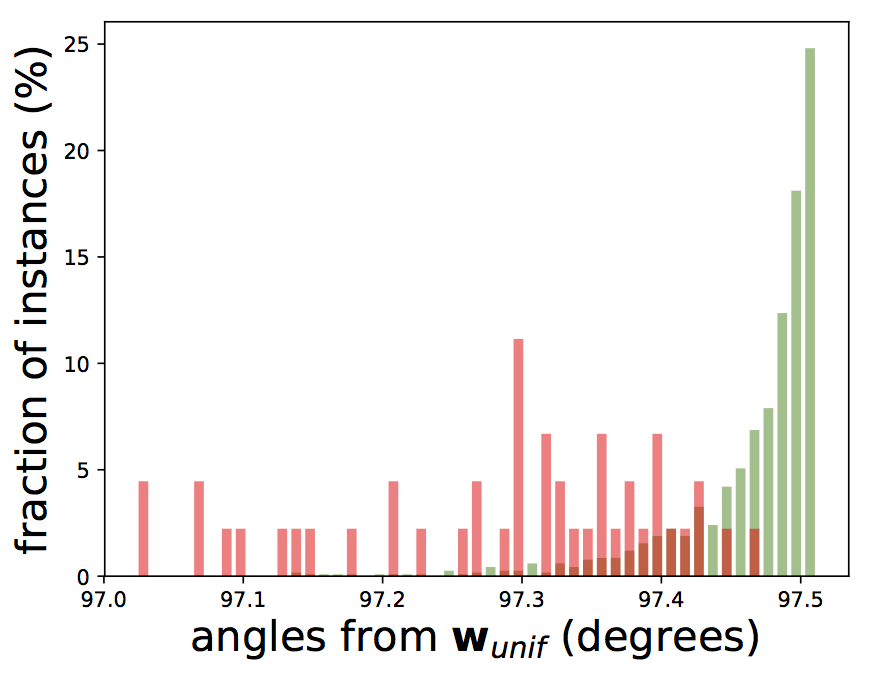}%
		\label{fig:angles_cardiotocography}}
	\subfloat[Covtype]{\includegraphics[width=0.24\textwidth]{angles/angles_covtype_iforest.png}%
		\label{fig:angles_covtype}} \\
	\subfloat[Electricity]{\includegraphics[width=0.24\textwidth]{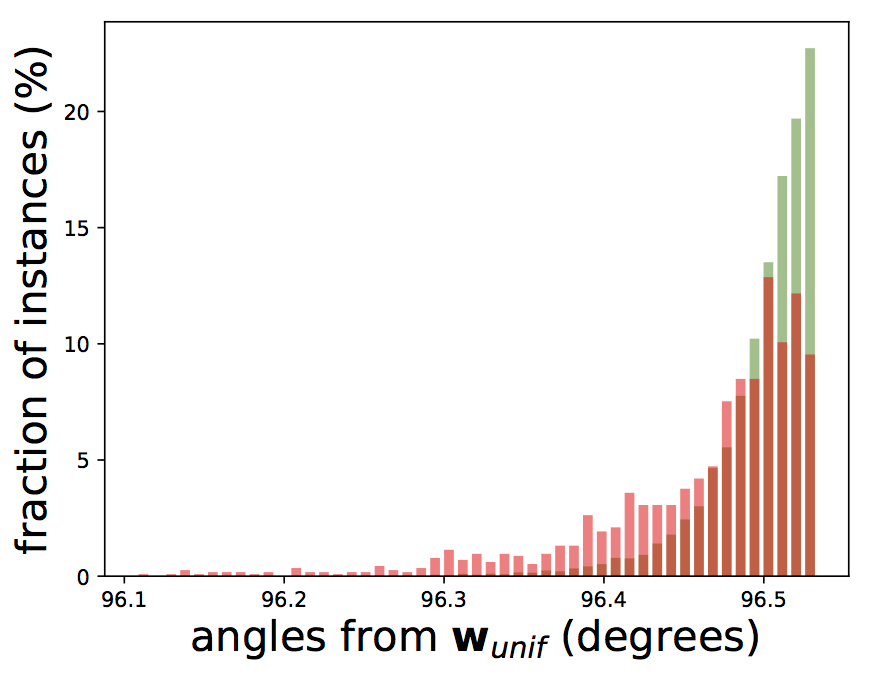}%
		\label{fig:angles_electricity}}
	\subfloat[KDDCup99]{\includegraphics[width=0.24\textwidth]{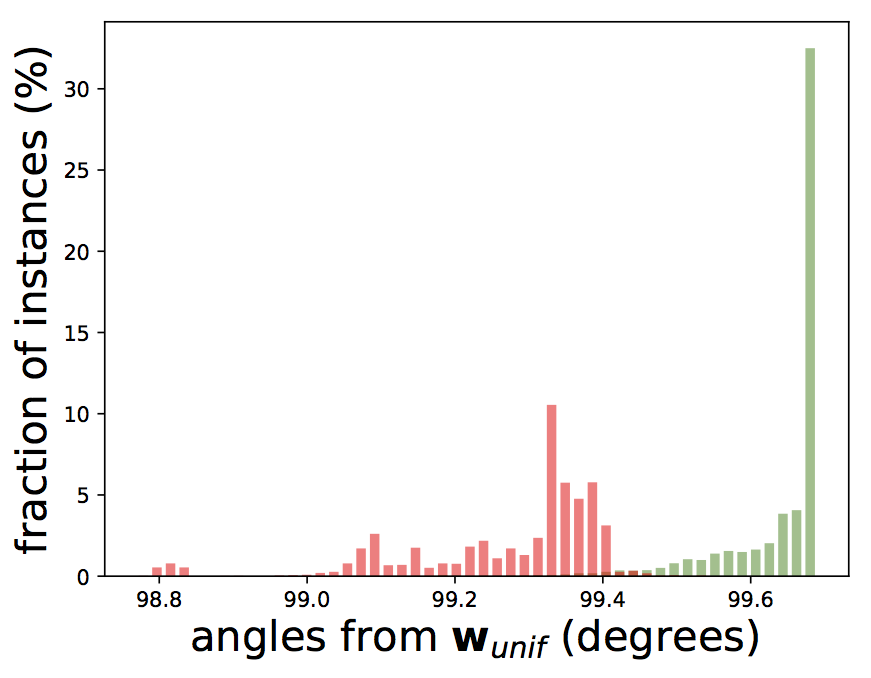}%
		\label{fig:angles_kddcup}}
	\subfloat[Mammography]{\includegraphics[width=0.24\textwidth]{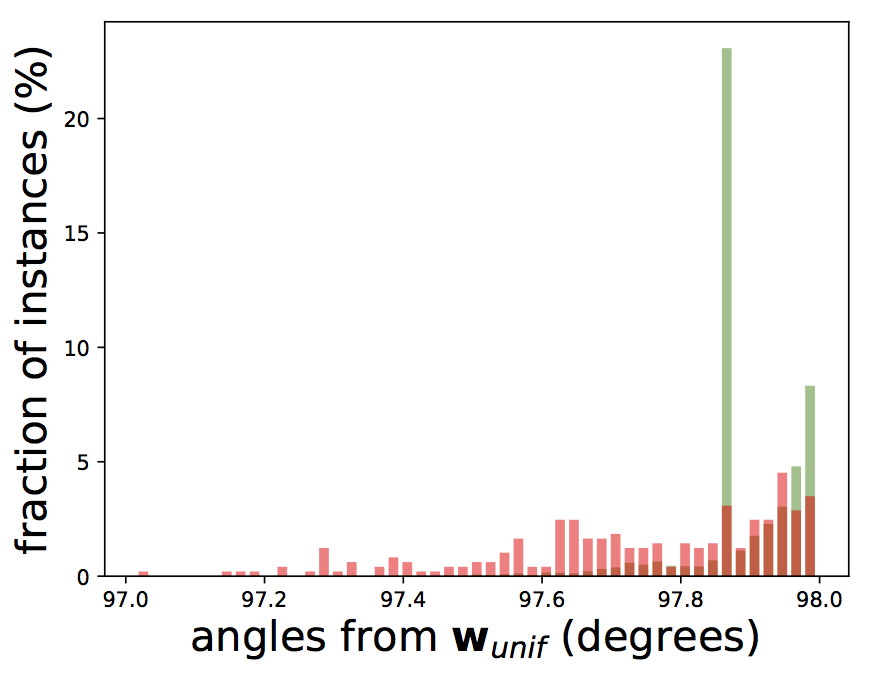}%
		\label{fig:angles_mammography}}
	\subfloat[Shuttle]{\includegraphics[width=0.24\textwidth]{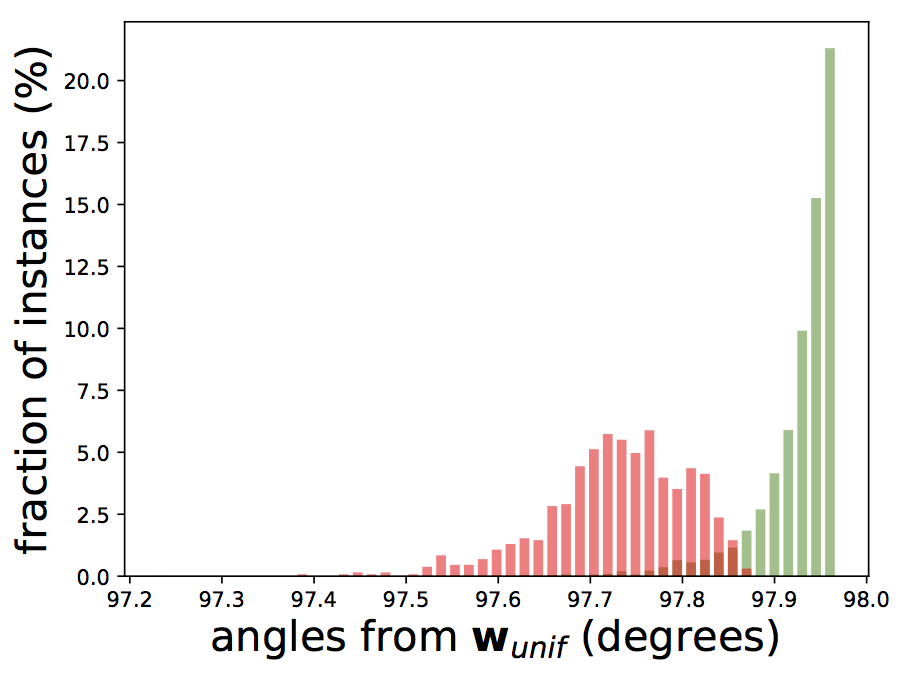}%
		\label{fig:angles_shuttle}} \\
	\subfloat[Weather]{\includegraphics[width=0.24\textwidth]{angles/angles_weather_iforest.png}%
		\label{fig:angles_weather}}
	\subfloat[Yeast]{\includegraphics[width=0.24\textwidth]{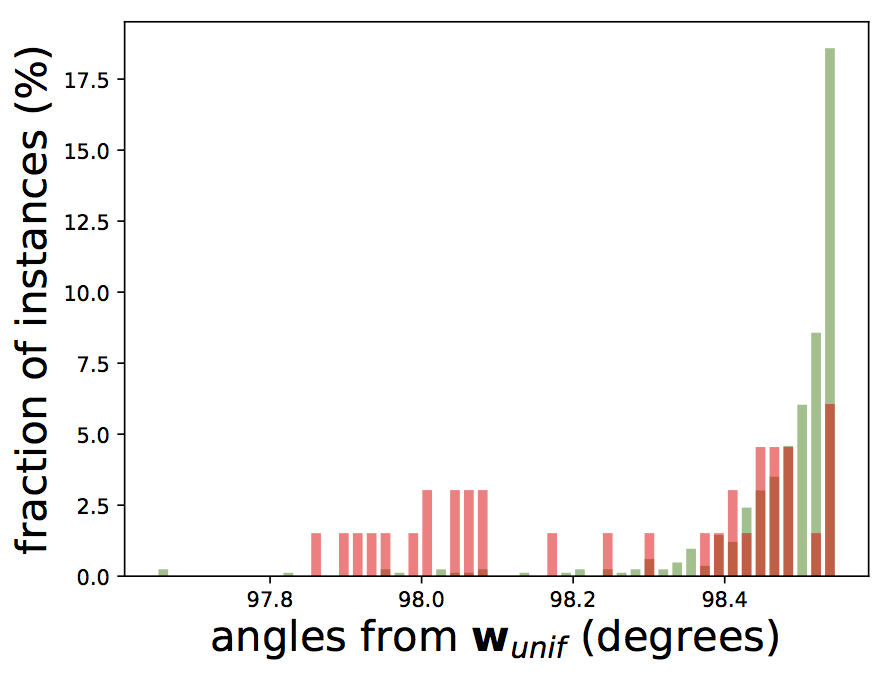}%
		\label{fig:angles_yeast}} \\[-1ex]
	\caption{Histogram distribution of the angles between score vectors from IFOR and ${\mathbf w}_{unif}$. The red and green histograms show the angle distributions for anomalies and nominals respectively. Since the red histograms are closer to the left, anomalies are aligned closer to ${\mathbf w}_{unif}$.}
	\label{fig:angles_all}
\end{figure*}

Table~\ref{tab:datasets_full} shows the complete details of the datasets used. Figure~\ref{fig:angles_all} shows that ${\mathbf w}_{unif}$ tends to have a smaller angular separation from the normalized IFOR score vectors of anomalies than from those of nominals. This holds true for most of our datasets (Table~\ref{tab:datasets_full}). Prior research has shown that \textit{Weather} and \textit{Electricity} are hard (Ditzler and Polikar 2013) and this is reflected in Figure~\ref{fig:angles_electricity} and Figure~\ref{fig:angles_weather}.
\captionsetup{belowskip=0pt,aboveskip=0pt}
\begin{table*}[h!]
	\captionsetup{font=large}
	\centering
	\footnotesize
	\caption{
		Description of datasets used in our experiments.}
	\renewcommand{\arraystretch}{0.4}
	{\begin{tabular}{llllll}
			\multicolumn{1}{l}{\bf Dataset} &\multicolumn{1}{l}{\bf Nominal Class} &\multicolumn{1}{l}{\bf Anomaly Class} &\multicolumn{1}{l}{\bf Total } & \multicolumn{1}{l}{\bf Dims } &\multicolumn{1}{l}{\bf \# Anomalies(\%)}
			\\ \hline \\
			Abalone         & 8, 9, 10 & 3, 21 & 1920 & 9 & 29 (1.5\%) \\
			\hline \\
			ANN-Thyroid-1v3         & 3 & 1 & 3251 & 21 & 73 (2.25\%) \\
			\hline \\
			Cardiotocography             & 1 (Normal) & 3 (Pathological) & 1700 & 22 & 45 (2.65\%) \\
			\hline \\
			Covtype             & 2 & 4 & 286048 & 54 & 2747 (0.9\%) \\
			\hline \\
			KDD-Cup-99             & \textit{`normal'} & \textit{`u2r', `probe'} & 63009 & 91 & 2416 (3.83\%) \\
			\hline \\
			Mammography             & -1 & +1 & 11183 & 6 & 260 (2.32\%) \\
			\hline \\
			Shuttle             & 1 & 2, 3, 5, 6, 7 & 12345 & 9 & 867 (7.02\%) \\
			\hline \\
			Yeast             & \textit{CYT}, \textit{NUC}, \textit{MIT} & \textit{ERL}, \textit{POX}, \textit{VAC} & 1191 & 8 & 55 (4.6\%) \\
			\hline \\ 
			Electricity             & \textit{DOWN} & \textit{UP} & 27447 & 13 & 1372 (5\%) \\
			\hline \\ 
			Weather             & \textit{No Rain} & \textit{Rain} & 13117 & 8 & 656 (5\%) \\
			\hline \\ 
	\end{tabular}}
	\label{tab:datasets_full}
\end{table*}

\subsection{Drift Detection for Datasets with Drift}
Each dataset has a streaming window size commensurate with its total size: \textit{Abalone}($512$), \textit{ANN-Thyroid-1v3}($512$), \textit{Cardiotocography}($512$), \textit{Covtype}($4096$), \textit{Electricity}($1024$), \textit{KDDCup99}($4096$), \textit{Mammography}($4096$), \textit{Shuttle}($4096$), \textit{Weather}($1024$), and \textit{Yeast}($512$).

\begin{figure*}
	\centering 
	\subfloat[Covtype]{
		\includegraphics[width=0.3\textwidth]{stream_diff/test_concept_drift_covtype.png}
		\label{fig:concept_drift_covtype_}}
	\subfloat[Electricity]{
		\includegraphics[width=0.3\textwidth]{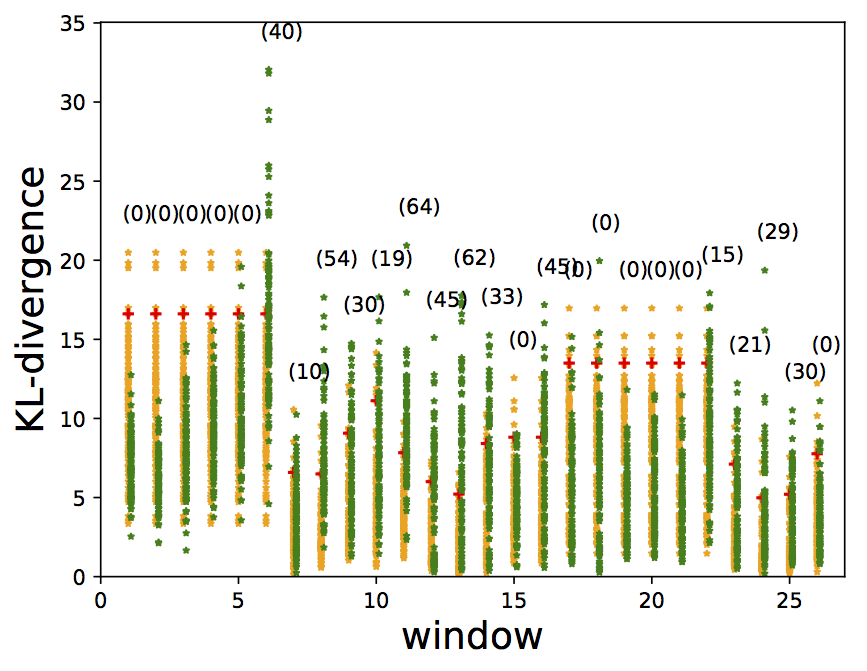}
		\label{fig:concept_drift_electricity_}}
	\subfloat[Weather]{
		\includegraphics[width=0.3\textwidth]{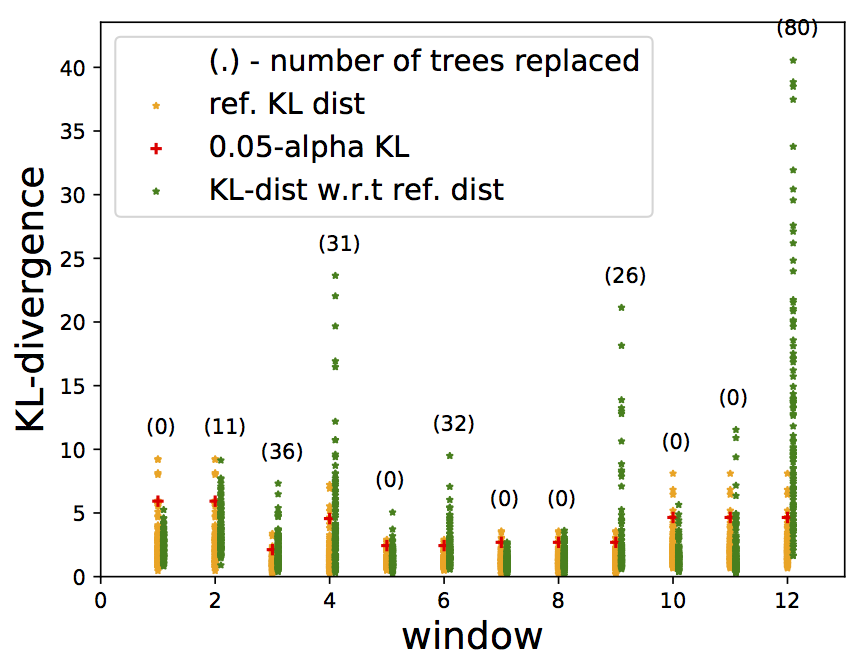}
		\label{fig:concept_drift_weather_}} \\
	\subfloat[Covtype]{
		\includegraphics[width=0.3\textwidth]{stream_diff/num_seen-covtype.png}
		\label{fig:concept_drift_covtype_num}}
	\subfloat[Electricity]{
		\includegraphics[width=0.3\textwidth]{stream_diff/num_seen-electricity.png}
		\label{fig:concept_drift_electricity_num}}
	\subfloat[Weather]{
		\includegraphics[width=0.3\textwidth]{stream_diff/num_seen-weather.png}
		\label{fig:concept_drift_weather_num}}
	\caption{Integrated drift detection and label feedback with Stream Active Learner (SAL). The top row shows the number of trees replaced per window when a drift in the data was detected relative to previous window(s). The bottom row shows the pct.~of total anomalies seen vs. number of queries for the \textbf{streaming datasets with significant concept drift}.}
	\label{fig:drift_detection_all_streaming}
\end{figure*}

Figure~\ref{fig:drift_detection_all_streaming} shows the results after integrating drift detection and label feedback with SAL for the datasets which are expected to have significant drift. Both \textit{Covtype} and \textit{Electricity} show more drift in the data than \textit{Weather} (top row in Figure~\ref{fig:drift_detection_all_streaming}). The rate of drift determines how fast the model should be updated. If we update the model too fast, then we loose valuable knowledge gained from feedback which is still applicable. Whereas, if we update the model too slowly, then the model continues to focus on the stale subspaces based on past feedback. It is hard to find a common rate of update that works well across all datasets (such as replacing $20\%$ trees with each new window of data -- \texttt{SAL~(Replace~20\%~Trees)} in bottom row of Figure~\ref{fig:drift_detection_all_streaming}). The adaptive strategy using KL-divergence (\texttt{SAL~(KL~Adaptive)}) (bottom row of Figure~\ref{fig:drift_detection_all_streaming}) seems to be robust and competitive with the best possible, i.e., \texttt{BAL}.

\begin{figure*}
	\centering 
	\subfloat[Abalone]{
		\includegraphics[width=0.3\textwidth]{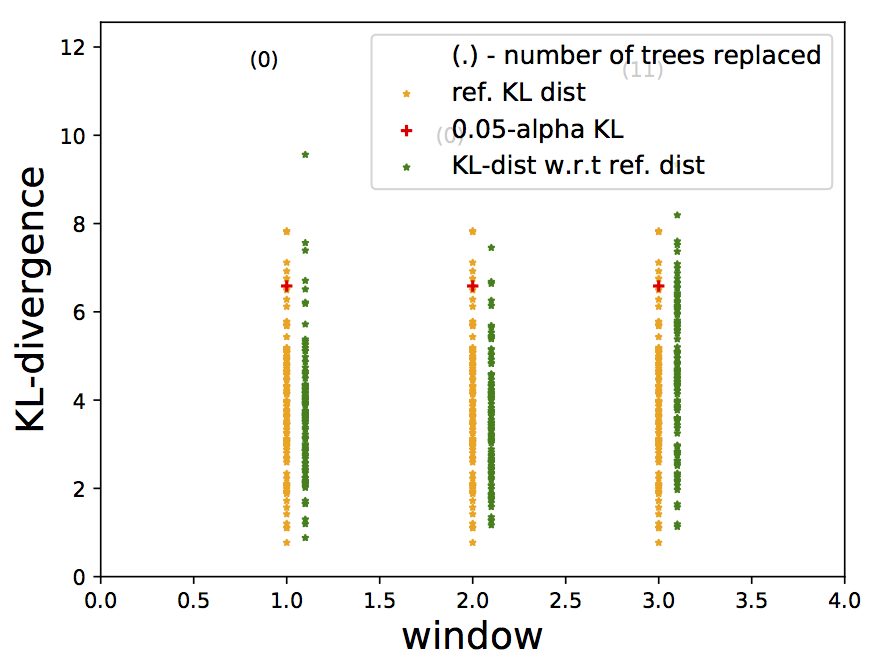}
		\label{fig:concept_drift_abalone_}}
	\subfloat[ANN-Thyroid-1v3]{
		\includegraphics[width=0.3\textwidth]{stream_diff/test_concept_drift_ann_thyroid_1v3.png}
		\label{fig:concept_drift_ann_thyroid_1v3_}}
	\subfloat[Cardiotocography]{
		\includegraphics[width=0.3\textwidth]{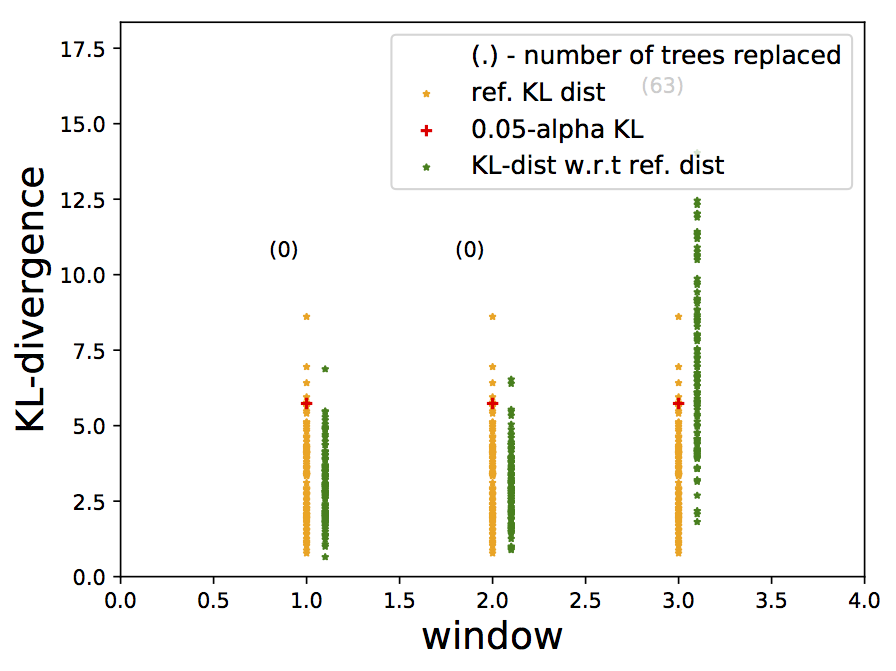}
		\label{fig:concept_drift_cardiotocography_1_}} \\
	\subfloat[Yeast]{
		\includegraphics[width=0.3\textwidth]{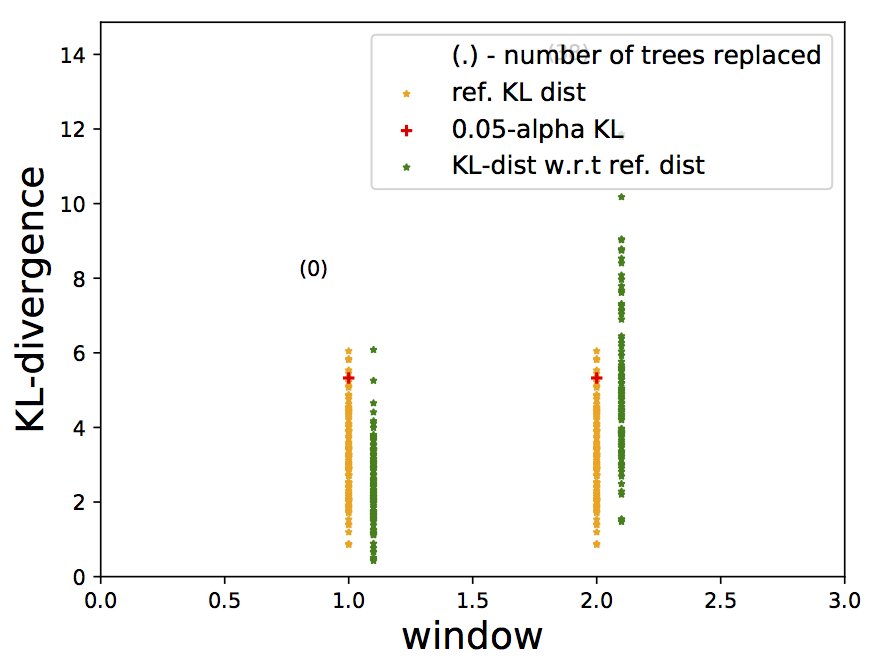}
		\label{fig:concept_drift_yeast_}}
	\subfloat[Mammography]{
		\includegraphics[width=0.3\textwidth]{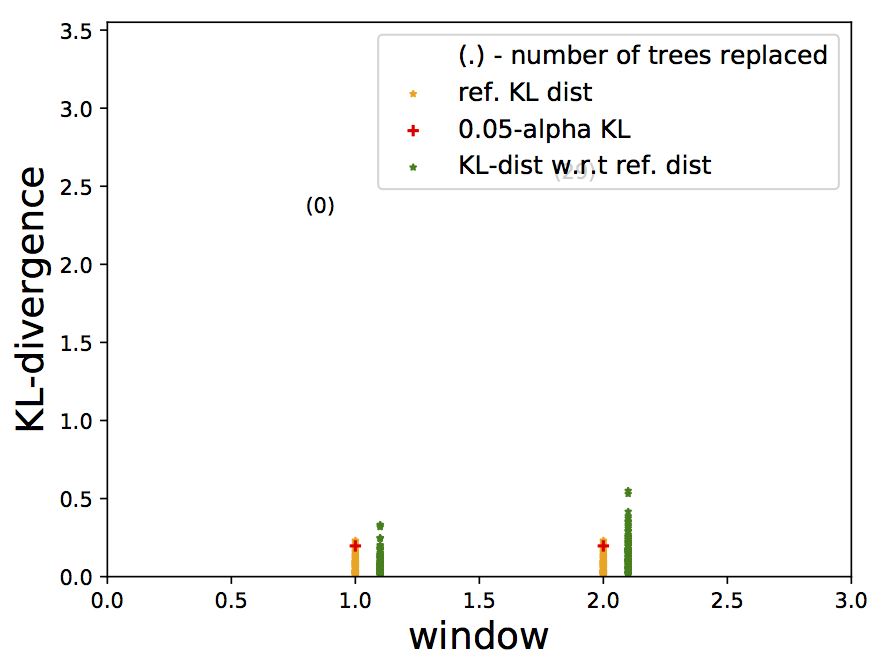}
		\label{fig:fig:concept_drift_mammography_}}
	\subfloat[KDD-Cup-99]{
		\includegraphics[width=0.3\textwidth]{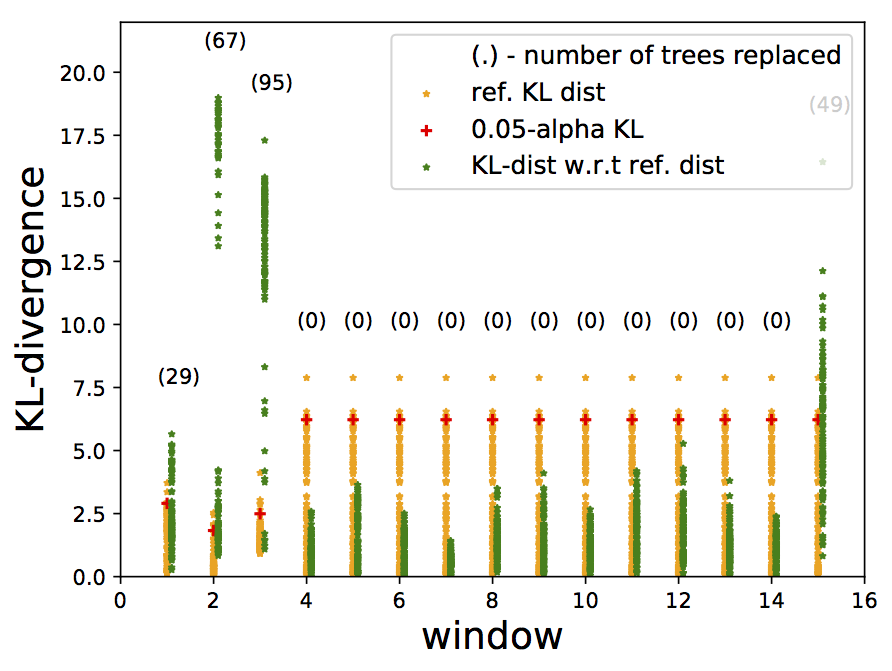}
		\label{fig:fig:concept_drift_kddcup_}} \\
	\subfloat[Shuttle]{
		\includegraphics[width=0.3\textwidth]{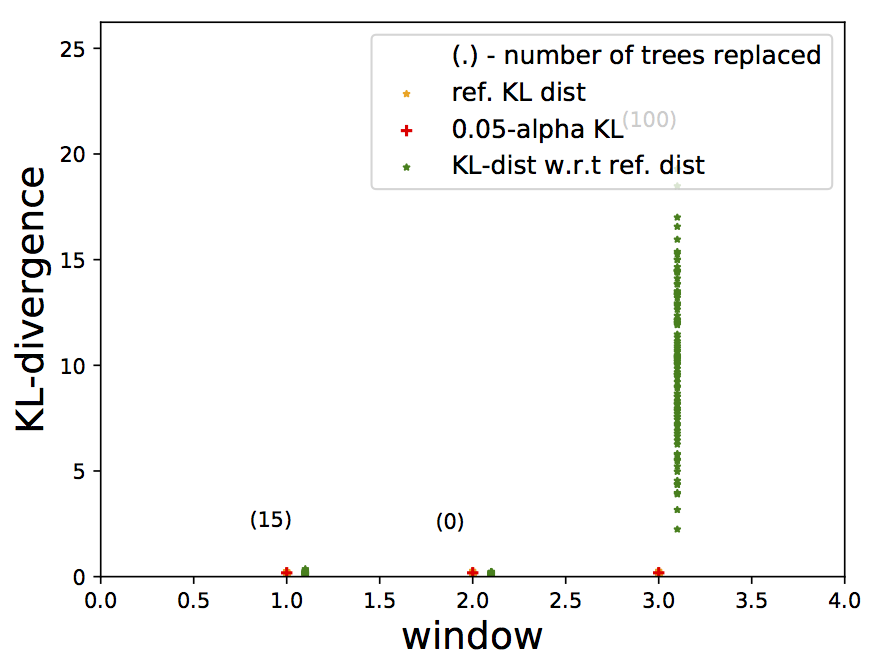}
		\label{fig:fig:concept_drift_shuttle_1v23567_}}
	\caption{The last data window in each dataset usually has much fewer instances and therefore its distribution is very different from the previous window despite there being no data drift. \textbf{Therefore, ignore the drift in the last window.} We did not expect \textit{Abalone}, \textit{ANN-Thyroid-1v3}, \textit{Cardiotocography}, \textit{KDDCup99}, \textit{Mammography}, \textit{Shuttle}, and \textit{Yeast} to have much drift in data, and this can also be seen in the plots where most of the windows in the middle of streaming did not result in too many trees being replaced (the numbers in the parenthesis are mostly zero).}
	\label{fig:drift_detection_all_non_streaming}
\end{figure*}

Figure~\ref{fig:drift_detection_all_non_streaming} shows the drift detection for the datasets which are not expected to have much drift.

\end{document}